\documentclass[fleqn,10pt]{wlscirep}
\usepackage[utf8]{inputenc}
\usepackage[T1]{fontenc}
\usepackage{hyperref}       
\usepackage{url}            
\usepackage{booktabs}       
\usepackage{amsfonts}       
\usepackage{nicefrac}       
\usepackage{microtype}      
\usepackage{xcolor}         
\usepackage{booktabs}
\usepackage{diagbox}
\usepackage{enumitem}
\usepackage{threeparttable}
\usepackage{xspace}
\usepackage{adjustbox}
\usepackage{amsmath}
\usepackage{wrapfig}
\usepackage{lineno}
\usepackage{cleveref}
\usepackage{caption}
\usepackage{longtable}
\usepackage{marvosym}
\usepackage{multirow}
\usepackage{longtable}
\usepackage{makecell}  
\usepackage[most]{tcolorbox}
\usepackage{etoolbox} 
\usepackage{setspace}
\usepackage{float}
\usepackage{afterpage}
\usepackage[section]{placeins} 

\crefname{figure}{Figure}{Figures}
\crefname{section}{Section}{Sections}
\crefname{table}{Table}{Tables}
\crefname{equation}{Equation}{Equations}
\crefname{appendix}{Appendix}{Appendixes}

\newcommand{\benchmark}{\textsc{CeProBench}\xspace}
\newcommand{\framework}{\textsc{CeProAgents}\xspace}
\newcommand{\knowledgegroup}{Knowledge Cohort\xspace}
\newcommand{\conceptgroup}{Concept Cohort\xspace}
\newcommand{\parametergroup}{Parameter Cohort\xspace}

\title{\framework: A Hierarchical Agents System for Automated Chemical Process Development}

\author[1*]{Yuhang Yang}
\author[1*]{Ruikang Li}
\author[2]{Jifei Ma}
\author[1,\Letter]{Kai Zhang}
\author[1]{Qi Liu}
\author[1]{Jianyu Han}
\author[1]{Yonggan Bu}
\author[2]{Jibin Zhou}
\author[1]{Defu Lian}
\author[1]{Xin Li}
\author[1,\Letter]{Enhong Chen}

\affil[1]{State Key Laboratory of Cognitive Intelligence, University of Science and Technology of China, Hefei, China}
\affil[2]{Dalian Institute of Chemical Physics, Chinese Academy of Sciences, Dalian, China}

\affil[*]{These authors contributed equally to this work}

\affil[\Letter]{kkzhang08@ustc.edu.cn, cheneh@ustc.edu.cn}

\begin{abstract}
The development of chemical processes, a cornerstone of chemical engineering, presents formidable challenges due to its multi-faceted nature, integrating specialized knowledge, conceptual design, and parametric simulation. 
Capitalizing on this, we propose \framework, a hierarchical multi-agent system designed to automate the development of chemical process through collaborative division of labor. Our architecture comprises three specialized agent cohorts focused on knowledge, concept, and parameter respectively. To effectively adapt to the inherent complexity of chemical tasks, each cohort employs a novel hybrid architecture that integrates dynamic agent chatgroups with structured agentic workflows. 
To rigorously evaluate the system, we establish \benchmark, a multi-dimensional benchmark structured around three core pillars of chemical engineering. We design six distinct types of tasks across these dimensions to holistically assess the comprehensive capabilities of the system in chemical process development.
The results not only confirm the effectiveness and superiority of our proposed approach but also reveal the transformative potential as well as the current boundaries of Large Language Models (LLMs) for industrial chemical engineering.

\end{abstract}
\begin{document}

\flushbottom
\maketitle
%
%
\thispagestyle{empty}

\section{Introduction}


The development of chemical processes is a core pillar of modern chemical engineering, underpinning a multitude of industries from pharmaceuticals and materials to energy and agriculture~\cite{pistikopoulos2021process, de2019would, burcham2018continuous, zimmerman2020designing, hassan2025membraneless}. However, understanding and developing these processes is profoundly challenging. This is an inherently multi-dimensional endeavor that demands the seamless integration of specialized domain knowledge, complex conceptual design, and rigorous parametric simulation~\cite{dimian2014integrated, grossmann2000research, luterbacher2025connecting}. For instance, engineers must navigate vast corpora of scientific literature and patents, conceptualize and validate complex process flow diagrams (PFDs), and perform detailed simulations to optimize parameters for a delicate balance of efficiency, safety, and economic viability~\cite{corriou2021chemical, schweidtmann2024mining}. This complex, iterative paradigm has traditionally relied heavily on human expertise, making it both time-consuming and resource-intensive~\cite{federsel2009chemical, he2023review}. To address this challenge, a promising approach is to leverage advanced artificial intelligence (AI) to forge a new paradigm of scientific intelligence for automated chemical process development~\cite{schweidtmann2024generative,zhou2025lab, quantrille2012artificial, li2025autonomous}.

Large Language Models (LLMs)~\cite{comanici2025gemini, yang2025qwen3, liu2025deepseek, jaech2024openai}, as powerful tools~\cite{xin2025towards}, have demonstrated profound capabilities for complex reasoning, knowledge synthesis, and task automation, significantly accelerating scientific discovery in fields such as chemistry\cite{li2026collective, mirza2025framework, llm-rdf2024}, biology\cite{wang2025geneagent, schaefer2025multimodal, tang2026deep}, and medical~\cite{wang2026making, tao2026llm, yang2026multimodal}.
Recently, the application of LLMs has extended into the domain of chemical engineering~\cite{llms4CE2025, srinivas2024accelerating, srinivas2025autochemschematic, gowaikar2024agentic, lee2024gpt, zeng2025llm}. For instance, to facilitate the initial design phase, an automated framework~\cite{srinivas2024accelerating} has been created to synthesize information from diverse online sources for designing process engineering schematics. Building on this, AutoChemSchematic AI~\cite{srinivas2025autochemschematic} presents a more comprehensive, physics-aware framework that automatically generates and validates industrially viable PFDs to bridge the gap from discovery to scale-up. Focusing on a key part of this design process, the ACPID Copilot~\cite{gowaikar2024agentic} utilizes a structured, multi-step workflow to automate the creation of PFDs directly from natural language descriptions. Beyond initial creation, the GIPHT~\cite{lee2024gpt} system employs a modular architecture that mimics the workflow of expert engineers to propose improvements to existing processes. Finally, for fine-tuning performance, the ProcessAgent~\cite{zeng2025llm} demonstrates that LLMs can autonomously infer operating constraints and guide the optimization of chemical processes when operating in a collaborative, structured manner.
 
Despite the significant progress in applying LLMs to chemical engineering, their current capabilities are hindered by several critical limitations that prevent them from addressing the holistic complexity of chemical process development~\cite{llms4CE2025}. 
Specifically, current LLM applications, while impressive in their specific niches, often struggle with the holistic, multi-faceted nature of chemical process development. The entire workflow demands not only complex reasoning and knowledge application but also seamless integration of sequential and parallel tasks across diverse domains. A single LLM or monolithic agent, despite its general intelligence, inherently lacks the specialized depth\cite{steyvers2025large}, robust tool integration\cite{ding2025scitoolagent}, and orchestrated collaboration\cite{su2025multi} required to master the entire pipeline from initial conceptualization to detailed parameter. This bottleneck restricts their ability to autonomously execute complex, interdependent engineering tasks that characterize real-world process design.
This fragmentation is evident in the current landscape: prior efforts have predominantly focused on isolated sub-tasks within the broader development cycle, such as initial schematic generation~\cite{srinivas2024accelerating, srinivas2025autochemschematic, gowaikar2024agentic} or parameter optimization~\cite{zeng2025llm}, rather than the entire process lifecycle. While valuable, these siloed approaches leave significant gaps, failing to provide a unified framework that can seamlessly integrate knowledge synthesis, conceptual design, and parameter simulation, which are interdependent and crucial for effective process development. Consequently, the transformative potential of AI for automated chemical process development in its entirety remains largely untapped.

In this study, we introduce \framework, a hierarchical and cooperative multi-agent system specifically engineered to automate and streamline these critical stages. 
Our architecture is designed to mimic the collaborative structure of a human engineering team and is organized into three distinct, specialized cohorts: a \knowledgegroup, a \conceptgroup, and a \parametergroup. This holistic organization seamlessly integrates these traditionally isolated stages, effectively resolving the fragmentation limitation of existing tools that focus on single sub-tasks.
Structurally, to master task complexity and ensure engineering rigor, each cohort integrates two parallel components: a ChatGroup~\cite{wu2024autogen, li2023camel, chatdev} for collaborative reasoning and planning, and a deterministic Workflow~\cite{hong2024metagpt, shen2023hugginggpt} for precise task execution. By synergizing flexible deliberation with standardized procedures, this hybrid architecture overcomes the reliability bottlenecks of monolithic models, ensuring both the creativity and precision required for complex engineering tasks.

To rigorously evaluate our system's capabilities and systematically benchmark the limits of current LLMs, we developed \benchmark, a comprehensive evaluation suite structured around three core engineering competencies: Knowledge, Concept, and Parameter.
This benchmark encompasses six distinct tasks designed to probe the depth of agentic intelligence across the full development lifecycle. In the Knowledge domain, we assess information processing capabilities through extraction and augmentation tasks. The Concept domain challenges agents with parsing, completion, and design tasks to evaluate topological reasoning and generation. Finally, the Parameter domain tests rigorous engineering fidelity through optimization tasks.  Our experimental results on \benchmark not only confirm the effectiveness of the \framework cooperative approach but also critically pinpoint the specific domains where LLMs currently fall short, providing a roadmap for future advancements in scientific AI.
\section{Results}


\begin{figure}[h!]
\centering
\includegraphics[width=\linewidth]{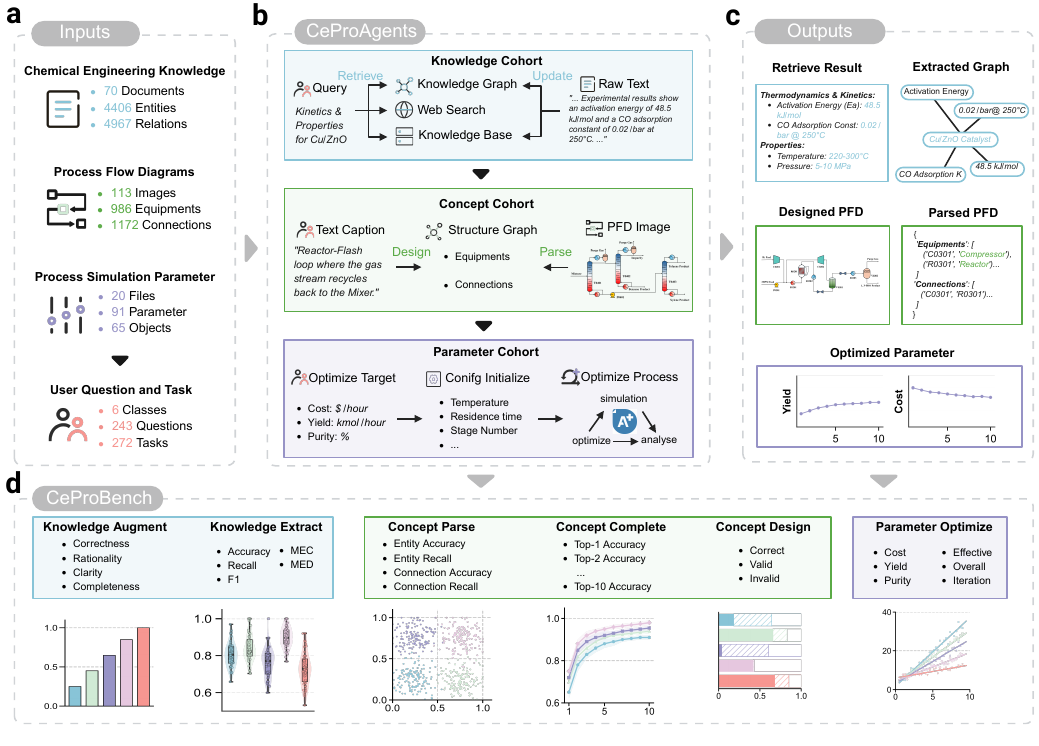}
\caption{\textbf{Overview of \framework and \benchmark.} \textbf{a,} The system processes heterogeneous chemical engineering inputs, including scientific documents, PFD imagery defining equipment and connections, and parameter files for simulation. \textbf{b,} The \framework architecture comprises three specialized cohorts: the \knowledgegroup retrieves chemical engineering knowledge from heterogeneous sources; the \conceptgroup bridges text captions, structural graphs, and PFD imagery; and the \parametergroup optimizes operational configurations for yield and economic targets. \textbf{c,} Representative multi-modal engineering results span three domains: retrieved kinetics and property data; designed and parsed PFD schematics; and optimized industrial parameters for yield and cost. \textbf{d,} The \benchmark suite probes six atomic tasks across the dimensions of Knowledge (Extract and Augment), Concept (Parse, Complete, and Design), and Parameter (Optimize).}
\label{fig:framework}
\end{figure}

\subsection{Overview of \framework and \benchmark}

The development of chemical processes involves a multi-dimensional integration of domain-specific knowledge, conceptual design, and rigorous parametric optimization (Figure \ref{fig:framework}a). To address the limitations of monolithic agents in managing these interdependent engineering tasks, we introduce \framework, a hierarchical multi-agent system designed to automate chemical process development through a collaborative division of labor. As illustrated in Figure \ref{fig:framework}b, the architecture is organized into three operational strata: the User Level for objective translation, the Cohort Level for collaborative task execution, and the Agent Level. This structure mirrors industrial engineering teams to ensure systematic handling of complex chemical engineering challenges.

The core of \framework comprises three specialized cohorts that bridge traditionally isolated stages of the development lifecycle. The \knowledgegroup serves as the epistemic foundation by executing synchronized retrieval protocols across curated internal databases\cite{chromadb_soft}, real-time web searches\cite{ddgs_soft}, and structured knowledge graphs\cite{neo4j_web}. This cohort facilitates the acquisition of specialized chemical engineering knowledge while mitigating inference hallucinations. Transitioning from theoretical data to practical design, the \conceptgroup acts as a multimodal reasoning engine that facilitates the interconversion between natural language descriptions, PFD imagery, and structured topological graphs. Through an adversarial collaboration environment, agents parse, propose, and critique design elements to ensure that generated blueprints remain topologically sound and compliant with chemical engineering principles. Finally, the \parametergroup functions as the validation and optimization engine. By interfacing with industry-standard computational engines such as Aspen Plus\cite{aspenplus_soft}, it implements an autonomous, closed-loop refinement cycle to iteratively optimize operating parameters including temperature, residence time, and stage numbers to satisfy multi-objective constraints regarding cost, yield, and purity. The representative outputs of these collaborative processes, ranging from structured knowledge graphs to optimized industrial parameters, are illustrated in Figure \ref{fig:framework}c.

To provide a rigorous quantitative assessment, we established \benchmark, a comprehensive evaluation suite designed to measure automated chemical engineering performance across six atomic tasks. These tasks span the three critical dimensions of Knowledge, Concept, and Parameter as outlined in Figure \ref{fig:framework}d. The Knowledge dimension evaluates informational fidelity through Knowledge Extract and Knowledge Augment tasks, requiring the system to process approximately 4,406 entities and 4,967 relations from 70 technical documents. The Concept dimension assesses the synthesis of process topologies via Concept Parse, Concept Complete, and Concept Design tasks, challenging the agent across 113 flow diagram scenarios with 986 equipments and 1172 connections. The Parameter dimension represents the most computationally intensive tier, centered on the Parameter Optimize task. This domain involves 20 specific simulation scenarios with 91 adjustable parameters, where performance is adjudicated by 65 objects. Using this multi-dimensional metric system, we systematically evaluated the \framework agents by instantiating them with five state-of-the-art foundation models: Gemini\cite{comanici2025gemini}, GPT\cite{lee2024gpt}, Claude\cite{claude}, Qwen\cite{yang2025qwen3}, and DeepSeek\cite{liu2025deepseek}. These experiments allow for the discernment of distinct reasoning capabilities and establish a robust baseline for chemical process intelligence.

\subsection{\framework unifies chemical engineering knowledge enhancement and extraction}

\begin{figure}[h!]
\centering
\includegraphics[width=1\linewidth]{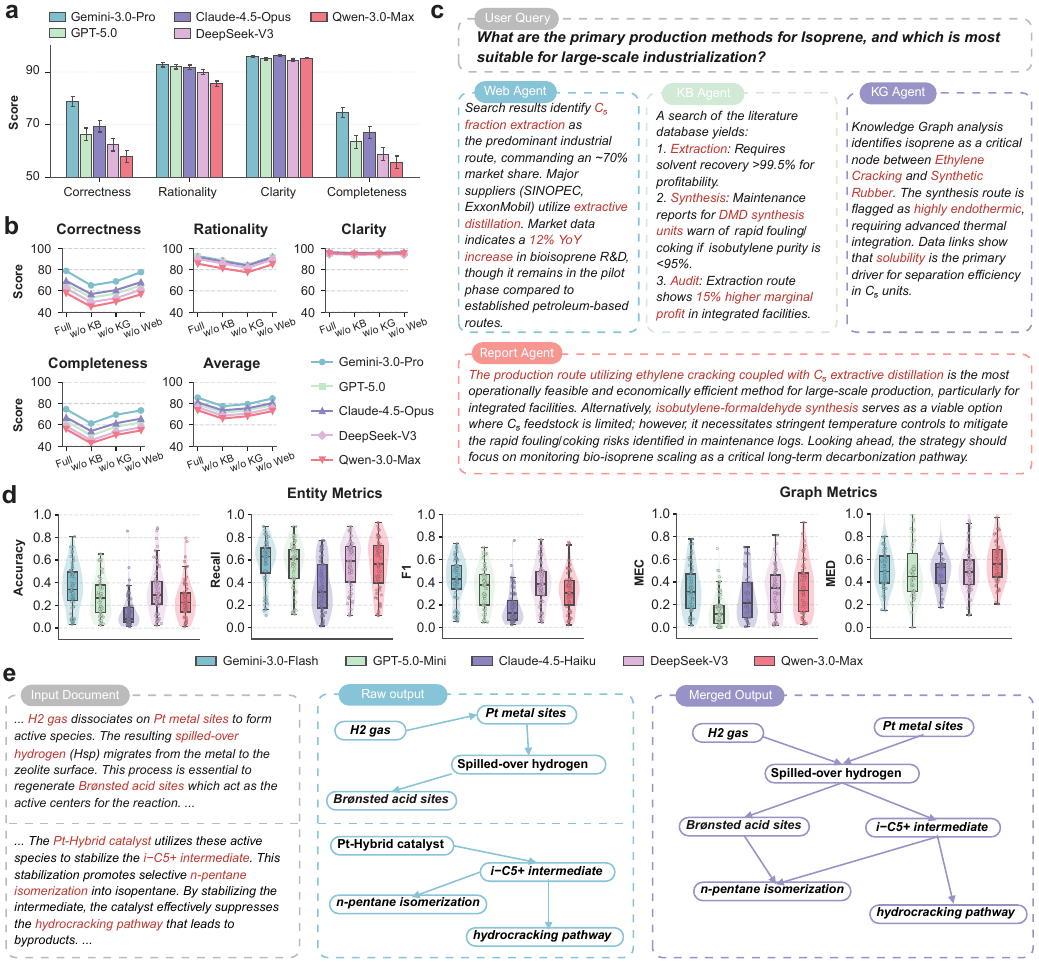}
\caption{\textbf{Chemical Knowledge Enhancement and Structured Extraction via \framework.} 
\textbf{a,} Quantitative assessment of knowledge enhancement performance across four dimensions: Correctness, Rationality, Clarity, and Completeness. 
\textbf{b,} Ablation studies demonstrating the impact of multi-source integration; scores are compared across Full configuration and versions excluding the Knowledge Base (w/o KB), Knowledge Graph (w/o KG), and Web Search (w/o Web). 
\textbf{c,} Case study of multi-agent collaborative retrieval for industrial production queries, illustrating the interplay between Web, KB, KG, and Report Agents. 
\textbf{d,} Statistical distribution of knowledge extraction performance, showing Entity Metrics (accuracy, recall, F1) and Graph Metrics (MEC, MED). 
\textbf{e,} Representative workflow of structured knowledge extraction, transitioning from unstructured academic text to a merged, disambiguated semantic graph.}
\label{fig:knowledges}
\end{figure}

The rigorous development of industrial-grade chemical processes necessitates the synthesis of heterogeneous external information and the structuring of internal domain knowledge. To address these challenges, the \knowledgegroup of \framework employs a collaborative architecture that unifies retrieval-augmented reasoning and structured graph construction.

\textbf{Synergistic Knowledge Enhancement.} To mitigate the hallucination risks inherent in standalone LLMs, \framework orchestrates a multi-agent group to collaboratively retrieve knowledge, significantly enhancing cognitive fidelity. As shown in Figure~\ref{fig:knowledges}a, the system consistently achieves high-fidelity performance across four critical dimensions comprising Correctness, Rationality, Clarity, and Completeness. In terms of Correctness, Gemini-3.0-Pro establishes a decisive lead, yielding a score of 78.8, which surpasses GPT-5.0 and Claude-4.5-Opus by margins of 12.5 and 9.4 points respectively, validating the efficacy of the framework in ensuring factual precision. For Rationality, the results exhibit a tiered distribution where the top-performing cohort, including Gemini-3.0-Pro, GPT-5.0, and Claude-4.5-Opus, clusters tightly above 91.0, whereas Qwen-3.0-Max trails at 85.5. While linguistic Clarity remains indistinguishable across all models with scores approaching the 96.0 ceiling, the integration of the \knowledgegroup reveals distinct performance tiers in Completeness, where Gemini-3.0-Pro outperforms DeepSeek-V3 and Qwen-3.0-Max by approximately 15.8 and 18.7 points respectively, highlighting the variance in information synthesis capabilities. 
The ablation study presented in Figure~\ref{fig:knowledges}b further substantiates the necessity of this multi-source integration. The removal of the Knowledge Base Agent results in a comprehensive degradation, causing Correctness and Completeness to plummet by 17.6 and 16.2 points respectively for the lead model in the absence of internal retrieval mechanisms. Crucially, the exclusion of the Knowledge Graph Agent leads to a distinct drop in Rationality across all models, reducing the Rationality score of Gemini-3.0-Pro from 92.7 to 81.2, which empirically confirms that structured ontological data is indispensable for maintaining logical consistency in complex chemical reasoning tasks.


The operational efficacy of this architecture is exemplified in the strategic analysis of Isoprene production shown in Figure~\ref{fig:knowledges}c. The Web Agent identifies SINOPEC as the dominant player in C5 fraction extraction, holding a 70\% market share, while also highlighting a 12\% rise in bio-isoprene R\&D activity. Complementing this, the Knowledge Base Agent mandates greater than 99.5\% solvent recovery for profitability and flags rapid fouling/coking if isobutylene purity drops below 95\%. Concurrently, the Knowledge Graph Agent characterizes the synthetic route as highly exothermic, necessitating advanced thermal integration. Integrating these streams, the Report Agent recommends extractive distillation due to a 15\% higher marginal profit, while proposing Isobutylene-Formaldehyde Synthesis as a conditional alternative for feedstock-limited scenarios. This synthesis elevates the system from simple retrieval to a context-aware strategy, mimicking human expert deliberation.


\textbf{Structured Knowledge Extraction.} To support and update the system's knowledge graph retrieval, our \framework implements a workflow that transforms unstructured technical literature into formalized knowledge graphs. We first evaluated the capacity for entity recognition using standard Entity Metrics as shown in Figure~\ref{fig:knowledges}d. The experimental results highlight the robust capabilities of Gemini-3.0-Flash, which secured the highest Entity F1 score of 0.431 driven by a Recall of 0.631 and an Accuracy of 0.340, validating its superior ability to identify discrete chemical concepts within dense academic texts. In comparison, DeepSeek-V3 and GPT-5.0-Mini achieved competitive F1 scores of 0.383 and 0.376 respectively, maintaining a balance between precision and coverage. Conversely, Qwen-3.0-Max and Claude-4.5-Haiku exhibited lower efficacy with F1 scores of 0.308 and 0.127 respectively, reflecting their limited capacity to accurately parse technical nomenclature from unstructured sources. Overall, these results confirm that state-of-the-art LLMs possess the requisite semantic precision to automate the digitization of domain-specific terminology, establishing a reliable nodal foundation for the subsequent knowledge graph construction.


Beyond isolated entities, the extraction of complex relationships proved more challenging across all backbones yet is essential for establishing the causal logic of chemical processes. We assessed this using Graph Metrics, specifically Mapping-based Edge Connectivity (MEC) and Mapping-based Edge Distance (MED). As illustrated in Figure~\ref{fig:knowledges}d, DeepSeek-V3 demonstrated superior performance in maintaining topological integrity, yielding a MEC score of 0.348, which indicates a robust grasp of relational dependencies. Qwen-3.0-Max and Gemini-3.0-Flash also maintained strong connectivity with scores of 0.324 and 0.313 respectively, ensuring that the extracted graphs remain structurally coherent with comparable Edge Distance values of approximately 0.49. This contrasts with GPT-5.0-Mini, which recorded a considerably lower connectivity score of 0.169, suggesting a specific limitation in capturing the interdependencies essential for process logic despite its reasonable performance in entity identification. Collectively, these structural evaluations underscore that capturing the intricate topological dependencies of chemical processes remains a distinctive challenge, necessitating models with advanced logical reasoning capabilities to ensure graph coherence.


We further elucidate the extraction performance through a representative workflow of the catalyst mechanism presented in Figure~\ref{fig:knowledges}e. The system processes raw text describing a Pt-Hybrid catalyst where the initial output identifies fragmented entities such as $H_2$ gas and Pt metal sites. Through the subsequent merging stage, the agent applies semantic reasoning to construct a coherent causal graph that links $H_2$ gas dissociation on Pt metal sites to the formation of spilled-over hydrogen, which subsequently migrates to regenerate Brønsted acid sites. This reconstruction captures the mechanism that the stabilization of the $i\text{-}C_5^+$ intermediate suppresses the hydrocracking pathway, thereby converting unstructured scientific literature into actionable engineering topologies suitable for downstream parameter initialization. Extended quantitative assessments and broader experimental validations are presented in Figure~\ref{fig:knowledge_Supplementary}.

\begin{figure}[h!]
\centering
\includegraphics[width=1\linewidth]{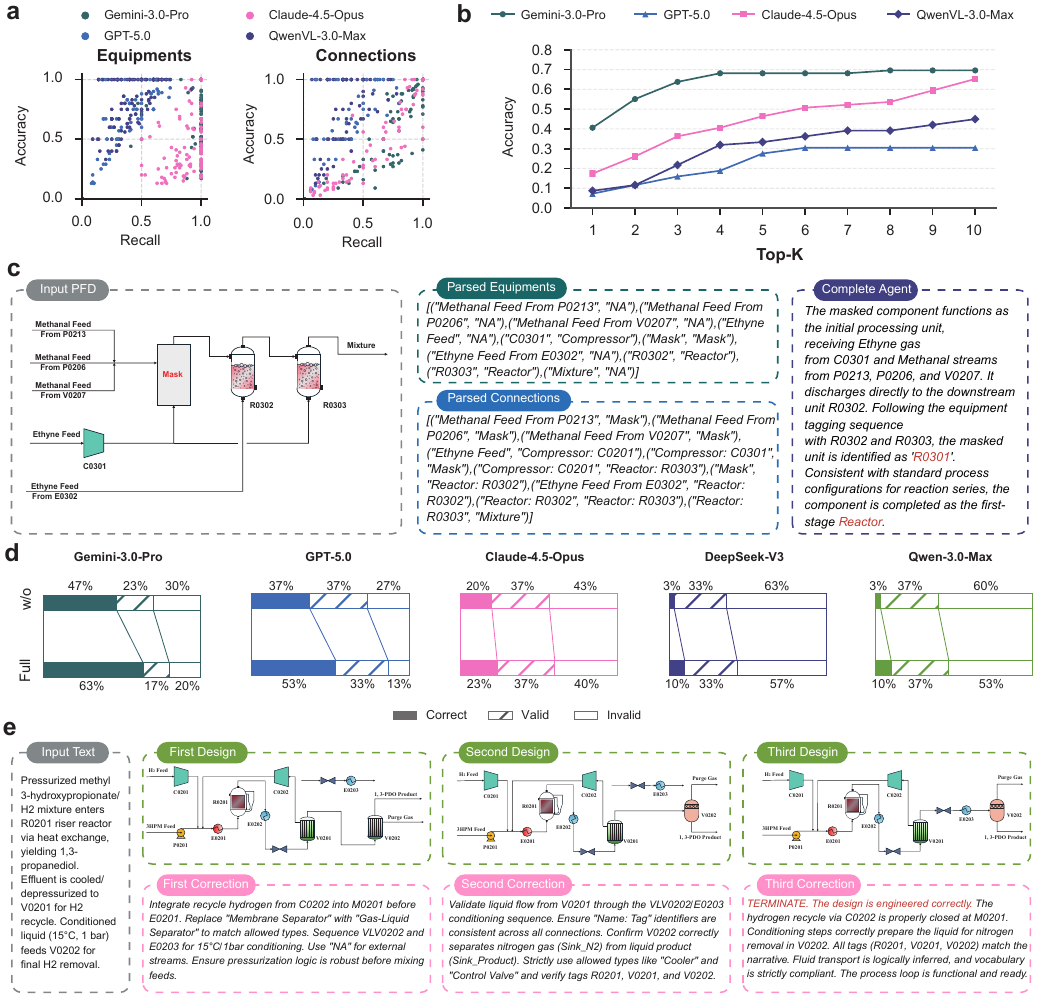}
\caption{\textbf{Process Concept Parsing, Completion, and Design via \framework.} 
\textbf{a,} Quantitative assessment of semantic parsing performance, showing accuracy and recall metrics for equipment and connection identification across different model backbones. 
\textbf{b,} Evaluation of generative completion capabilities using Top-K accuracy metrics, highlighting the system's proficiency in inferring missing process components compared to baseline models. 
\textbf{c,} Representative case study illustrating the complete workflow from semantic parsing of raw PFDs to the logical completion of underspecified topologies. 
\textbf{d,} Evaluation of generative design performance comparing the Full configuration against an ablation without the Correction Agent (w/o).
\textbf{e,} Detailed case study of iterative generative design, demonstrating how the system refines initial proposals through a multi-turn correction loop to satisfy complex engineering constraints.}
\label{fig:concept}
\vspace{-0.25cm}
\end{figure}

\subsection{\framework automates process diagram parsing, completion, and generative design}

The \conceptgroup of \framework establishes a collaborative engineering framework capable of automated diagrammatic reasoning. As shown in Figure~\ref{fig:concept}, this cohort effectively closes the loop between static visual data and dynamic process logic.


\textbf{High-Fidelity Parsing and Topological Completion.} The fundamental prerequisite for automated process development is the high-fidelity digitization of static schematics into semantic graphs. As quantified by the scatter plots in Figure~\ref{fig:concept}a, we systematically evaluated the parsing fidelity of \framework when instantiated with various backbones. In the equipment identification task, Gemini-3.0-Pro establishes a dominant baseline, achieving a superior balance with both accuracy (72\%) and recall (88\%) leading the cohort. In contrast, GPT-5.0 exhibits a conservative retrieval strategy; while it sacrifices recall to approximately 37\%, its resulting accuracy (68\%) still fails to surpass that of Gemini. Transitioning to the more demanding connectivity parsing task, the performance gap widens further. Gemini-3.0-Pro remains the only model to maintain robustness with both connection precision and recall exceeding 65\%. Conversely, GPT-5.0 experiences a substantial performance drop with accuracy falling to 47\%, failing to maintain the structural coherence required for reliable autonomous parsing despite retaining a slight edge over lower baselines.


Beyond structural digitization, the system's cognitive depth is rigorously assessed through the generative completion of underspecified topologies, a task that demands deep domain intuition to infer missing unit operations. Figure~\ref{fig:concept}b presents the quantitative evaluation using Top-K accuracy metrics, revealing distinct reasoning trajectories across model backbones. \framework, when instantiated with Gemini-3.0-Pro, exhibits superior inferential precision; its performance curve is characterized by a sharp initial ascent, achieving a leading Top-1 accuracy of approximately 40.6\% and rapidly converging to a saturation point of 68.1\% within the top 5 candidates. This trajectory indicates a decisive ability to capture implicit engineering rules without ambiguity. In contrast, Claude-4.5-Opus displays a steady, linear growth pattern, while both GPT-5.0 and QwenVL-3.0-Max plateau early at significantly lower accuracy levels (below 45\%). These results not only validate the necessity of a reasoning-strong backbone but, more importantly, demonstrate the efficacy of the \framework architecture in successfully grounding these foundation models within the rigorous constraints of chemical engineering, enabling them to solve complex topological dependencies that were previously intractable.


This statistical superiority is underpinned by a sophisticated cognitive workflow, vividly detailed in the representative case study of formaldehyde ethynylation presented in Figure~\ref{fig:concept}c. The process initiates with the parsing of a raw, unstructured PFD, where the system must navigate visual noise to extract a semantic graph. As illustrated, the Equipment Parsing Agent accurately digitizes the upstream acetylene feed train, correctly identifying the compressor (C0301) and resolving the intricate web of methanal feed lines (from P0213 and P0206). However, the critical challenge lies in the masked component, an underspecified node positioned centrally within the reaction series. Instead of relying on superficial pattern matching, the Completion Agent employs deep engineering deduction. By analyzing the downstream topology, the agent identifies that the subsequent unit, R0302, is explicitly tagged as a ``Reactor'' within a multi-stage ethynylation train. Leveraging domain knowledge regarding cascade reactor configurations for this synthesis pathway, the agent logically infers that the masked unit must function as the primary reactor (R0301) of the identical equipment class. This successful recovery of the ``Reactor'' identity, derived solely from topological context, exemplifies the system's capacity to transcend simple visual recognition and engage in causal engineering reasoning.


\textbf{Generative Design and Adversarial Correction.} Progressing from analysis to synthesis, the system is further tasked with the generative design of industrial-grade PFDs. Figure~\ref{fig:concept}d evaluates this core competency, highlighting both the effectiveness of our generative framework and the critical necessity of the adversarial correction mechanism. 
In terms of design efficacy, the framework, when powered by reasoning-strong backbones like Gemini-3.0-Pro, successfully synthesizes fully correct engineering blueprints in 63\% of cases under the full configuration, demonstrating a robust grasp of process logic. 
However, the ablation study reveals that relying solely on one-pass generation without correction limits performance to 47\%, often resulting in legal but non-functional schematics shown as striped regions. The activation of the correction loop is therefore decisive, as it not only propels the correctness rate by 16 percentage points but also suppresses logically incorrect outputs from 30\% to 20\%. 
A similar trajectory is observed in GPT-5.0, where the framework elevates design fidelity from 37\% to 53\%, with invalid outputs dropping to a system-low of 13\%. Even for models with lower baselines like DeepSeek-V3 and Qwen-3.0-Max, the framework enhances correctness from 3\% to 10\%. 
These results confirm that while the Design Agent provides the creative synthesis, the multi-agent Correction loop acts as the essential engineering filter, ensuring the final flowcharts are not just visually coherent but operationally viable.

This iterative refinement capability is vividly detailed in the case study presented in Figure~\ref{fig:concept}e regarding the continuous production of 1,3-propanediol. In the initial translation of the textual description, the Design Agent correctly identified the riser fluidized bed reactor (R0201) but committed a vocabulary violation by proposing a ``Membrane Separator'', a term absent from the strict equipment ontology. The Correction Agent, functioning as an engineering auditor, immediately flagged this discrepancy and simultaneously identified a topological flaw: the hydrogen recycle loop via C0202 was not properly integrated into the mixing point M0201. Through the subsequent Design-Critique interaction, the system autonomously replaced the invalid unit with a standard ``Gas-Liquid Separator'' and rectified the recycle logic. A final validation loop further optimized the downstream conditioning sequence to ensure precise nitrogen removal in V0202 ($15^{\circ}\text{C}, 1 \text{bar}$), allowing the system to converge on a final design that is not only topologically valid but also operationally compliant with the specific process narrative. Extended quantitative assessments and broader experimental validations are presented in Figure~\ref{fig:concept_Supplementary}.

\begin{figure}[p]
\centering
\includegraphics[width=1\linewidth]{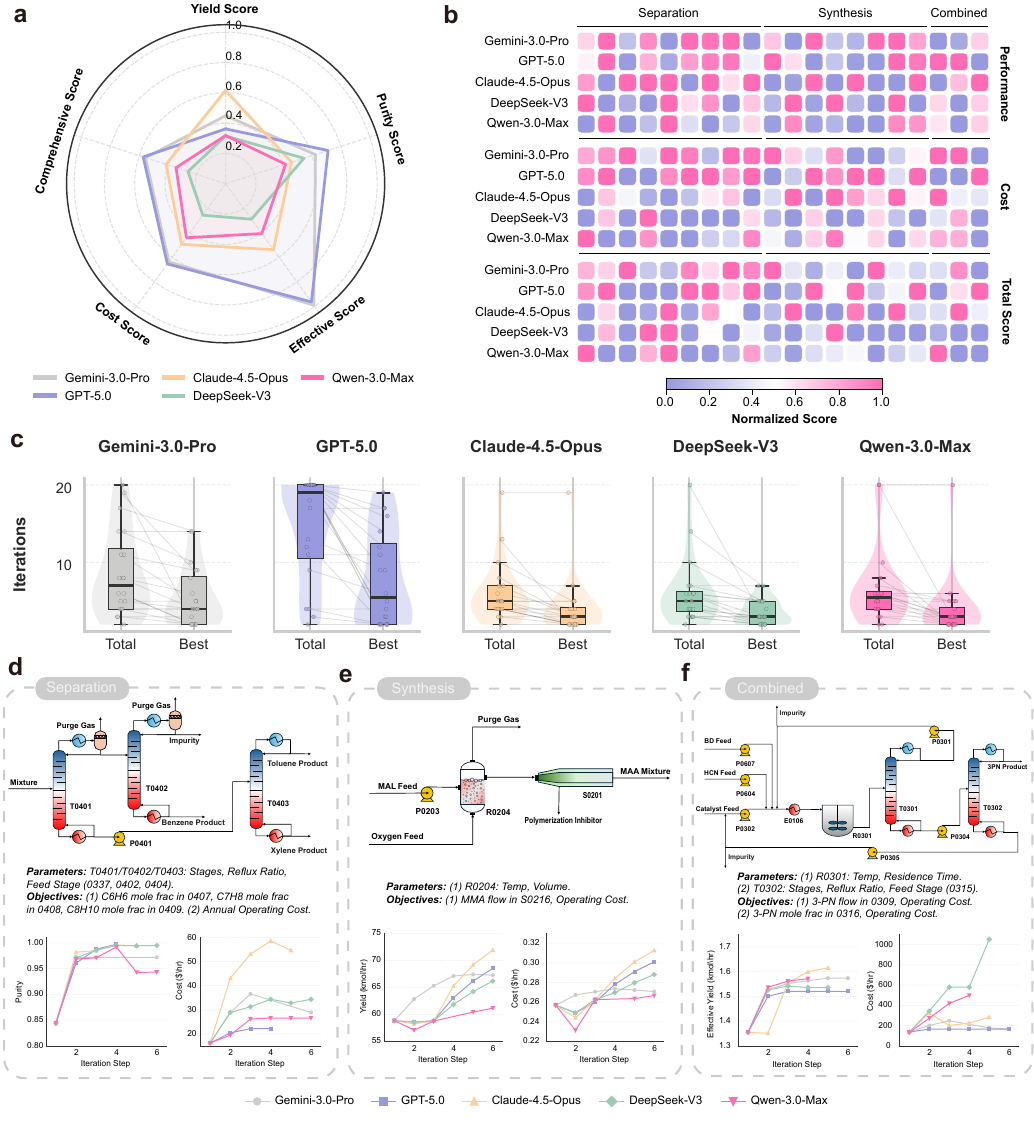}
\caption{\textbf{Industrial Process Parameter Optimization via \framework.} 
\textbf{a,} Quantitative assessment of multi-objective optimization performance. The radar chart illustrates the normalized scores across five critical dimensions including Cost, Yield, Purity, and Iteration efficiency.
\textbf{b,} Comparative analysis of optimization efficacy stratified by distinct unit operation types (Separation, Synthesis, and Combined).
\textbf{c,} Evaluation of convergence efficiency illustrating the distribution of total optimization iterations and the specific iteration steps where the global optimum was identified. The box plots coupled with scatter points highlight the search speed and decision-making decisiveness of different backbones.
\textbf{d-f,} Case studies of autonomous parameter optimization across distinct chemical processes.
\textbf{d,} Separation process for aromatics purification involving a multi-column distillation sequence.
\textbf{e,} Synthesis process for MMA production.
\textbf{f,} Combined process for butadiene hydrocyanation to produce 3-PN.}
\label{fig:parameter}
\end{figure}
\subsection{\framework streamlines and accelerates industrial-grade process parameter optimization}

Moving beyond structural synthesis, the \parametergroup addresses the complexity of dynamic process validation. By integrating autonomous agents with industrial-grade simulators, this module iteratively refines operating conditions within a rigorous physicochemical framework. This ensures that the generated designs translate into economically and technically feasible operations, satisfying multi-dimensional engineering constraints.

\textbf{Holistic Optimization Performance.} 
To rigorously quantify the automated engineering capabilities in high-dimensional search spaces, we evaluated the system by aggregating the normalized performance across five critical dimensions: Comprehensive Score, Cost Score, Effective Score, Purity Score, and Yield Score (Figure \ref{fig:parameter}a). \framework, particularly when instantiated with Gemini-3.0-Pro, achieved a dominant performance profile by maximizing the Effective Score to approximately 0.99 and establishing a state-of-the-art baseline. Meanwhile, GPT-5.0 demonstrated a highly competitive profile by excelling in Purity Score (0.71) and Cost Score (0.65). This reflects its robust ability to optimize economic viability while strictly adhering to quality constraints. Notably, Claude-4.5-Opus secured the highest Yield Score of 0.62 although its performance across other metrics remained more balanced. In contrast, models such as DeepSeek-V3 and Qwen-3.0-Max exhibited visible contractions particularly in the Comprehensive Score and Cost Score dimensions. For instance, DeepSeek-V3 recorded a Cost Score of only 0.26 and a Comprehensive Score of 0.26 which significantly trailed the leading models. This divergence highlights a critical limitation in weaker backbones as they struggle to navigate the non-convex optimization landscape to find globally optimal solutions that satisfy multi-dimensional engineering constraints.
Beyond aggregate metrics, we stratified performance across distinct unit operations including separation, synthesis, and combined tasks (Figure \ref{fig:parameter}b). In synthesis tasks, the performance gap narrowed as most models achieved normalized scores above 0.6. This indicates a generalized proficiency in kinetic maximization across different backbones. However, distinct capability tiers emerged in separation and combined scenarios. Reasoning-strong backbones like Gemini-3.0-Pro and GPT-5.0 maintained robustness with scores exceeding 0.8 as they successfully managed the non-linear thermodynamics inherent in processes like azeotropic distillation. Conversely, models lacking deep logical consistency struggled in these complex domains and often yielded scores below 0.3.

\textbf{Convergence Efficiency.} 
The operational efficiency of the agentic workflow is quantified by contrasting the total optimization iterations versus the iteration steps required to achieve the best results (Figure \ref{fig:parameter}c). This comparison reveals three distinct search and decision-making patterns. Gemini-3.0-Pro exhibits the most ideal strategic termination behavior. Median data indicates that it typically identifies the global optimum by the 4th iteration and autonomously determines convergence to cease operations around the 7th iteration. This compact gap between total and best iterations demonstrates the high decisiveness of the model which minimizes computational costs while ensuring performance. In contrast, GPT-5.0 displays a clear tendency towards over-exploration. Although it identifies optimal solutions rapidly with a median best iteration of 5.5, its total iteration count reaches a median of 19. This implies that it often exhausts the iteration budget to verify existing results which leads to redundant consumption. Conversely, models such as Claude-4.5-Opus, DeepSeek-V3, and Qwen-3.0-Max record median total iterations of only 5 to 5.5 and cease performance growth by the 3rd iteration. When viewed alongside their lower comprehensive scores in the holistic assessment, this extremely short search cycle does not indicate efficiency but rather reflects premature convergence or insufficient search depth where the models terminate tasks before fully exploring the solution space.

\textbf{Autonomous Optimization Case Study.} 
To demonstrate the operational logic of \framework in real-world engineering scenarios, we present three representative cases across three process types: separation, synthesis, and combined processes.

In the Separation case (Figure \ref{fig:parameter}d), the system was tasked with the sharp separation of a complex aromatics mixture. The Gemini-3.0-Pro instantiated agents exhibited a decisive search trajectory, rapidly satisfying the purity constraint by reaching a plateau of 97.2\% in just the second iteration. Once the quality threshold was secured, the Internal Analyst shifted the objective to economic minimization. This strategic pivot successfully reduced the operating cost from a peak of 36.6 \$/hr in Iteration 3 to 28.9 \$/hr by Iteration 6, a substantial 21\% reduction, demonstrating the system's capability to navigate the trade-off between product quality and operational expense.

For the Synthesis task (Figure \ref{fig:parameter}e) involving Methyl Methacrylate (MMA) production, the optimization profiles reveal distinct reasoning behaviors. While Claude-4.5-Opus achieved the highest terminal yield of 71.9 kmol/hr, it incurred a disproportionately high operating cost of 0.313 \$/hr. In contrast, \framework instantiated with Gemini-3.0-Pro identified a superior Pareto-optimal configuration (Yield: 67.3 kmol/hr, Cost: 0.271 \$/hr). By autonomously balancing reaction kinetics with utility consumption, the system avoided the diminishing returns of over-intensification observed in other baselines.

The most complex scenario, Combined process optimization for duce 3-pentenenitrile (3-PN) production (Figure \ref{fig:parameter}f), exemplifies the robust hierarchical synergy of the \parametergroup. The system initially prioritized Effective Yield, boosting it from 1.35 to a saturation point of approximately 1.57 kmol/hr within the first three iterations. Crucially, the subsequent phase highlighted the system's stability: whereas models like DeepSeek-V3 experienced severe instability with costs spiking to over 1000 \$/hr, the Gemini-driven agents autonomously executed a cost-minimization strategy. This drove annual operating expenses down by 33\% (from a peak of 248 \$/hr to 165 \$/hr) in the final optimization stages, achieving a high-fidelity convergence that satisfies both engineering and economic criteria. Extended quantitative assessments and broader experimental validations are presented in Figure~\ref{fig:parameter_Supplementary}.

\begin{figure}[t]
\centering
\includegraphics[width=\linewidth]{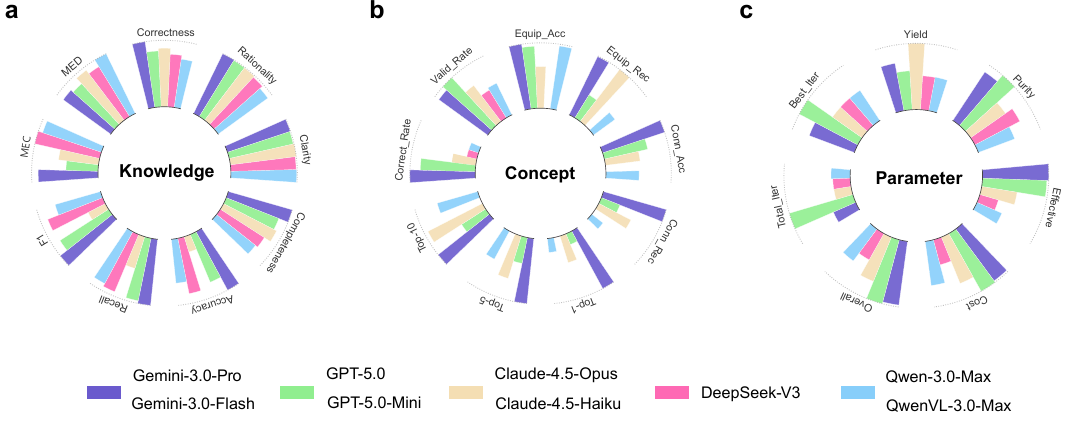}
\caption{\textbf{Overall performance assessment of \framework on \benchmark.} 
\textbf{a,} The Knowledge dimension evaluation comparing epistemic fidelity and structural extraction capabilities. The radar chart aggregates metrics for reasoning quality (Correctness, Rationality, Clarity, Completeness) and graph construction integrity (Accuracy, Recall, F1, MEC, MED). 
\textbf{b,} The Concept dimension evaluation assessing multimodal topological reasoning across three distinct tasks: semantic parsing (Equipment/Connection Accuracy and Recall), topological completion (Top-K Accuracy), and generative design (Valid Rate and Correct Rate). 
\textbf{c,} The Parameter dimension evaluation quantifying parameter optimization efficacy. The chart contrasts optimization outcomes (Yield, Purity, Effective, Cost, Overall) with computational efficiency (Total vs. Best Iterations) and the aggregate Effective Score.}
\label{fig:results}
\end{figure}

\subsection{\framework demonstrates robust holistic performance across knowledge, concept, and parameter}

To provide a holistic assessment of the \framework framework, we conducted a comprehensive comparative analysis across the three critical pillars of chemical engineering: Knowledge, Concept, and Parameter. Figure \ref{fig:results} synthesizes these multi-dimensional results, elucidating the performance ceilings and distinct reasoning behaviors of five LLMs when instantiated within our agentic architecture.

In the Knowledge dimension (Figure \ref{fig:results}a), the system demonstrates that high-fidelity epistemic reasoning is contingent upon a synergy between the agentic workflow and the backbone model's intrinsic capability. Gemini-3.0-Pro and GPT-5.0 exhibit a dominant performance profile, forming a comprehensive envelope across both reasoning metrics (Correctness, Rationality, Clarity) and graph extraction metrics (MEC, MED). While weaker backbones often maintain competitive linguistic Clarity, they exhibit a discernible contraction in structural metrics such as MEC and F1-score. This indicates that while the \framework architecture can enforce logical consistency, the extraction of complex, long-range causal dependencies in chemical literature remains a differentiator for reasoning-strong models.

Transitioning to the Concept dimension (Figure \ref{fig:results}b), the performance divergence becomes most pronounced in tasks requiring multimodal topological grounding. The radar chart reveals that Gemini-3.0-Pro achieves the most balanced performance across all three sub-tasks. In semantic parsing, it leads in the digitization of raw schematics, maximizing both Equipment and Connection Accuracy. In topological completion, the high Top-K accuracy scores demonstrate its superior intuition in inferring missing engineering components from underspecified contexts. Crucially, in generative design, performance is adjudicated by Valid Rate and Correct Rate. While most models achieve a high Valid Rate, indicating they can generate structurally coherent and connected graphs, the Correct Rate reveals a significant divergence. Leading models distinguish themselves by minimizing engineering violations, ensuring that the generated designs are not just topologically legal but operationally functional.

Finally, the Parameter dimension (Figure \ref{fig:results}c) illustrates the trade-off between optimization quality and computational efficiency. The Overall and Effective scores are maximized by Gemini-3.0-Pro and GPT-5.0, which consistently locate the Pareto front for Yield and Purity while managing Cost. The Iteration axes (Total vs. Best) provide critical insight into the agents' decision-making decisiveness. While some models show low Total Iteration counts, their corresponding Overall scores are suboptimal, suggesting premature convergence. Conversely, the leading models demonstrate an optimal search strategy-expending sufficient computational steps to maximize the Effective Score without engaging in redundant exploration, thereby achieving a superior balance between engineering performance and computational cost.
\section{Discussion}

The traditional development of chemical processes has long relied on the intricate, cross-disciplinary collaboration of human experts, spanning the full lifecycle from multi-source literature synthesis to PFDs design and complex parameter simulation~\cite{dimian2014integrated, federsel2009chemical}. However, existing monolithic LLMs encounter significant cognitive bottlenecks when addressing such tasks, which are characterized by strict physical constraints, high reliability requirements, and tightly coupled logic~\cite{comanici2025gemini, jaech2024openai}. Consequently, current AI approaches struggle to achieve the systemic cognitive integration necessary to unify the fragmented stages of process development, spanning from heterogeneous literature synthesis to rigorous parameter simulation~\cite{llms4CE2025}. This systemic integration deficit represents a fundamental barrier, severely restricting the practical application and reliability of LLMs in the chemical engineering domain, thereby hindering their ability to empower industrial-grade process development.

To transcend these limitations, we introduce \framework, a hierarchical and highly collaborative multi-agent system designed to automate the holistic chemical process development workflow. The architecture is structured around three specialized cohorts: the \knowledgegroup, which constructs a dynamic epistemic foundation; the \conceptgroup, acting as a multimodal reasoning engine for topological generation; and the \parametergroup, which executes closed-loop parameter optimization. By synergizing divergent ChatGroups for strategic planning with convergent Workflows for precise task execution, \framework effectively reconciles generative creativity with the deterministic rigor required in engineering design.

To rigorously quantify automated engineering capabilities, we established \benchmark, a multi-dimensional evaluation suite designed around the three core pillars of chemical engineering. Collectively, these domains constitute a comprehensive dataset comprising 6 task classes, spanning a total of 243 questions and 235 specific tasks. Specifically, the testing environment encompasses the Knowledge domain, processing 4,406 entities and 4,967 relations extracted from 70 technical documents. In the Concept domain, it includes 113 PFDs featuring 986 equipment units and 1,172 connections. Furthermore, for the Parameter domain, the benchmark incorporates 20 high-fidelity parameter files with 91 adjustable parameters and 65 optimization objectives, where performance is adjudicated by stringent engineering metrics, thereby establishing a robust standard for chemical process intelligence.

Empirical results substantiate the efficacy of \framework in bridging the critical gap between generative reasoning and rigorous engineering requirements. Our multi-dimensional analysis reveals that the hierarchical agentic architecture significantly outperforms monolithic baselines by strictly enforcing structural consistency and operational viability. In the knowledge domain, the synergistic multi-source retrieval mechanism effectively mitigated epistemic hallucinations, elevating the factual correctness of scientific reasoning to a leading score of 78.8. In the conceptual domain, the deployment of the adversarial correction mechanism proved decisive, elevating the fidelity of engineering blueprints to a peak of 63\% and effectively suppressing the invalid topologies that typically hinder unguided generation. Furthermore, in the parameter domain, the agent-driven closed-loop optimization demonstrated a superior capacity to navigate high-dimensional trade-offs, realizing a 40\% reduction in operating costs while maximizing process yield. These quantitative breakthroughs confirm that synergizing structured workflows with LLMs enables the autonomous and reliable resolution of high-dimensional industrial challenges.

While \framework establishes a transformative paradigm for AI-driven chemical engineering, future research must address two critical frontiers. First, the depth of the system's current logical deduction remains bounded by the reasoning floor of the underlying models; future work must explore embedding thermodynamic laws and physical constraints more explicitly into the agents' decision-making chains to enhance industrial-grade precision in complex systems. Second, integrating the \framework architecture deeply with the physical world, through the incorporation of real-time experimental feedback and industrial control systems, to participate directly in realistic chemical production scenarios. This convergence of digital intelligence and physical operation~\cite{li2025autonomous, zhou2025lab}, which will be a critical pathway for moving beyond simulation to achieve actual industrial deployment, sustainable manufacturing, and green hydrogen production~\cite{de2019would, zimmerman2020designing, nogueroles2025polyethylene}.
\section{Methods}

This section details the methodologies underpinning our work, including the architecture of \framework and the construction of the benchmark \benchmark for its rigorous assessment.

\subsection{\framework Architecture}
\framework is structured as a vertically integrated hierarchy designed to harmonize high-level conceptual design with rigorous technical execution. This orchestration originates at the system’s apex, where the User defines overarching mission objectives. These objectives are systematically decomposed into distinct constituent parts and propagated into an intermediate stratum of collective intelligence.
This layer is composed of specialized Functional Cohorts that undertake and collaborate on these segmented tasks, emulating the multi-disciplinary deliberation and collaborative dynamics of a professional engineering environment. This process facilitates the transition from abstract conceptualization to structured design logic, which is eventually operationalized at the system's granular foundation.
Here, Low-level Agents serve as operative primitives, leveraging a diverse computational toolkit to execute discrete technical tasks through deterministic workflows. By partitioning the architecture into these interconnected tiers, \framework achieves a synergistic integration of top-down strategic reasoning and robust bottom-up engineering validation.

\textbf{The Cohort Architecture.} At the foundational level, we define an individual autonomous agent $a_i$ as a functional primitive that encapsulates its cognitive state and operative transformation:
\begin{equation}
a_i = \langle m_i, o_i, t_i, x_i, y_i \rangle, \quad f(m_i, o_i, t_i) : x_i \to y_i,
\end{equation}
where $m_i$ denotes the core LLM serving as the neural processor, $o_i$ represents the objective function specified via system prompt, and $t_i$ refers to the suite of accessible computational tools. The mapping $f$ characterizes the agent's internal reasoning process, whereby input $x_i$ is translated into a output $y_i$ within the context of its defined model and task-specific objectives.
To catalyze collective intelligence and high-level deliberation, agents are organized into a ChatGroup $G$, formulated as a collaborative reasoning structure:
\begin{equation}
G = \langle A_G, C \rangle, \quad A_G = \{a_1, \dots, a_m\}, \quad C = \{c_i \mid c_i \subseteq A_G\}.
\end{equation}
The ChatGroup facilitates information exchange via communication channels $C$, allowing agents in $A_G$ to resolve complex, multi-objective conceptual challenges through multi-turn, discursive interactions.
In parallel, a complementary execution structure is instantiated as a Workflow $F$, designed for sequential task-solving and validated engineering outputs:
\begin{equation}
F = \langle A_F, E \rangle, \quad A_F = \{a_{m+1}, \dots, a_{m+n}\}, \quad R = \{r_{ij} = \langle a_i, a_j \rangle \mid a_i, a_j \in A_F, x_j = y_i\}.
\end{equation}
The Workflow structure governs the deterministic flow of data through a pipeline $R$, where the output $y_i$ of a precursor agent is precisely mapped to the input $x_j$ of its successor, ensuring rigorous state transitions throughout the chemical process design.
The structural synthesis of deliberative reasoning and algorithmic execution is achieved through the Cohort $S$:
\begin{equation}
S = \{G, F, I\}, \quad I = \{i_{ij} = \langle a_i, a_j \rangle \mid a_i \in A_G, a_j \in A_F, x_j = y_i\}.
\end{equation}
In this unified architecture, $I$ serves as the critical instruction bridge, representing the inter-modular translation of intelligence. Within the Cohort, the high-level cognitive planning emerged from $G$ generates executable directives to steer the operational Workflow $F$, thereby ensuring that abstract strategic objectives are systematically operationalized into robust, validated implementations.

\textbf{The \knowledgegroup.} 
Functioning as the framework's epistemic substrate, the \knowledgegroup $S_{knowledge}$ unifies real-time context augmentation with continuous memory consolidation to sustain cognitive fidelity. This integration is achieved by coordinating the ChatGroup $G_{knowledge}$, which actively retrieves multi-modal evidence to mitigate inference hallucination, with the Workflow $F_{knowledge}$, designed to systematically assimilate unstructured data into the system's internal repositories.
To mitigate information sparsity, the ChatGroup $G_{knowledge}$ orchestrates a synchronized retrieval protocol. Upon receiving a query $q$, the agents $a_{web}$, $a_{kb}$, and $a_{kg}$ concurrently activate distinct cognitive pathways to harvest constituent knowledge streams:
\begin{equation}
k_{web} = \textsc{WebSearch}(q_{web}), \quad k_{kb} = \textsc{VectorQuery}(q_{kb}), \quad k_{kg} = \textsc{GraphQuery}(q_{kg}),
\end{equation}
where $q_{web}, q_{kb}, q_{kg}$ denote the reformulated query, vector embedding, and extracted entities, respectively. To resolve redundancy and factual discrepancies among these heterogeneous sources, a Report Agent $a_{report}$ dialectically integrates the raw streams to produce a coherent context $k_{re} = \textsc{Report}(k_{web}, k_{kb}, k_{kg})$, serving as a hallucination-free basis for subsequent reasoning.
The systematic ingestion of unstructured data is governed by the Workflow $F_{knowledge}$. Given an input document $D$, the pipeline initially partitions it into semantic chunks $D = \{D_1, D_2, \dots, D_n\}$. Each chunk $D_i$ undergoes dual processing: it is encoded by an embedding model for vector storage, and simultaneously processed by Knowledge Extraction Agent $a_{extract}$ to instantiate a local subgraph $KG_i = \langle V_i, E_i \rangle$. Here, $V_i$ denotes the set of identified entities, and $E_i = \{\langle s_j, p_j, o_j \rangle \mid s_j, o_j \in V_i\}$ represents the extracted relational triples. 
These local structures are subsequently aggregated into a unified global graph $KG = \bigcup_{i} KG_i$. To ensure ontological consistency within $KG$, we implement a semantic disambiguation protocol. For any entity pair $(v_i, v_j)$, we calculate the semantic similarity $s_{ij}$ based on embedding proximity. The resolution logic follows a tri-state threshold mechanism:
\begin{equation}
\textsc{Resolve}(v_i, v_j) = 
\begin{cases} 
\textsc{AutoMerge} & \text{if } s_{ij} > t_{auto} \\
\textsc{ReviewMerge} & \text{if } t_{review} < s_{ij} \le t_{auto} \\
\textsc{Distinct} & \text{if } s_{ij} \le t_{review}.
\end{cases}
\end{equation}
Pairs falling into the critical review interval trigger the Merge Agent $a_{merge}$ to adjudicate identity based on contextual evidence, balancing automation efficiency with the high precision required for scientific rigor.

\textbf{The \conceptgroup.} 
Functioning as the multimodal reasoning engine of \framework, the \conceptgroup $S_{concept}$ establishes a unified transformation among natural language descriptions, PFD imagery, and structured topological graphs.
The digitization of static PFDs is governed by the Topology Parsing Workflow $F_{concept}$. The Parser Agent $a_{parser}$ orchestrates a serialized and dependency-aware execution protocol.
The process initiates with the Equipment Parsing Agent $a_{equip}$, which is empowered by a multimodal LLM to map pixel regions into standardized unit operation ontologies $V$.
Subsequently, the Link Parsing Agent $a_{link}$ operates under a context-injection mechanism. It receives the identified node set $V$ as a prior condition, strictly constraining its search space to flow lines $E$ that originate and terminate at valid equipment ports. This conditional extraction ensures the resulting topology $G_{raw}$ is free from floating edges, mathematically modeled as:
\begin{equation}
V = \textsc{EquipExtract}(\text{PFD}), \quad E = \textsc{LinkExtract}(\text{PFD} \mid V), \quad G_{raw} = \textsc{Combine}(V, E),
\end{equation}
where $\textsc{LinkExtract}$ denotes that edge perception is structurally subordinated to the recognized entity set.
In the domain of conceptual design, the synthesis of engineering topologies from natural language objectives is facilitated by the Concept Design ChatGroup $G_{concept}$. This group functions as an adversarial collaboration environment that harmonizes creative design with rigorous engineering validation.
The Design Agent $a_{design}$ serves as the structural architect, proposing an initial graph layout based on heuristic logic. Conversely, the Correction Agent $a_{cor}$ acts as a senior process auditor. It evaluates the proposed topology against a knowledge base of Chemical Engineering Logic $K_{chem}$.
This interaction constitutes a dialectic optimization loop. Let $G^{(t)}$ be the graph state at iteration $t$. The refinement process is decoupled into critical auditing and structural updating:
\begin{equation}
\begin{aligned}
O^{(t)} = \textsc{Correct}(G^{(t)}, K_{chem}), \quad G^{(t+1)} = \textsc{Design}(G^{(t)}, O^{(t)}),
\end{aligned}
\end{equation}
where $O^{(t)}$ represents the natural language critique identifying operational hazards generated by the auditor, and $\textsc{Refine}$ denotes the generative update applied by $a_{des}$ conditioned on both the prior state and the provided critique.
The Completion Agent $a_{comp}$ serves as the ultimate integration layer to guarantee the integrity of cross-modal transformations. Addressing the information sparsity inherent in isolated artifacts, $a_{comp}$ contextualizes the graph within the broader project scope. It infers necessary structural augmentations to ensure the final output $G_{final}$ is logically complete, distinct, and ready for downstream simulation tasks.

\textbf{The \parametergroup.}
To bridge the gap between conceptual process topologies and rigorous physical realization, we instantiate the \parametergroup $S_{parameter}$. This cohort functions as the system's validation and optimization engine, responsible for transforming Abstract Graphs into high-fidelity engineering models and iteratively refining operational parameters to satisfy user-defined performance metrics such as yield, cost, and safety.
The initialization phase is governed by the Parameter Strategy ChatGroup $G_{parameter}$, which functions as a multi-disciplinary collaborative environment. Its primary objective is to analyze the provided flow diagram and translate abstract user requirements into a machine-readable configuration file and precise parameter initialization ranges (JSON).
To ensure simulation viability, $G_{parameter}$ orchestrates four specialized experts: the Sentinel Expert ($a_{sent}$) analyzes the PFD topology to define critical state and control variables; the Chemist Expert ($a_{chem}$) provides thermodynamic insights into reaction kinetics; and the Inspector Expert ($a_{insp}$) establishes hard safety limits and quality specifications.
Finally, the Initialize Expert ($a_{init}$) functions as the strategic decision-maker, synthesizing these multi-domain inputs to define the optimal search space $\Omega$ and initial boundary conditions, mathematically formalized as the initialization vector $P^{(0)}$.
The execution of multi-objective optimization is managed by the Parameter Optimization Workflow $F_{parameter}$, which implements an autonomous, agent-driven closed-loop refinement cycle. This workflow interfaces with external industry-standard solvers (e.g., Aspen Plus) and operates through the iterative synergy of three internal agents.
Let $P^{(t)}$ be the parameter set at iteration $t$. The cycle proceeds as follows:
First, the Internal Optimizer ($a_{opt}$) generates candidate parameters within the defined search space.
Second, the Internal Simulator ($a_{parameter}$) executes the high-fidelity simulation primitives to obtain performance metrics $R^{(t)}$.
Third, the Internal Analyst ($a_{ana}$) evaluates these outputs against historical trends to generate diagnostic feedback $A^{(t)}$.
This iterative process continues until convergence criteria are met, modeled as:
\begin{equation}
\begin{aligned}
R^{(t)} = \textsc{Simulate}(P^{(t)}), \quad
A^{(t)} = \textsc{Analyze}(R^{(t)}, \text{History}), \quad
P^{(t+1)} = \textsc{Optimize}(P^{(t)}, A^{(t)}),
\end{aligned}
\end{equation}
where $\textsc{Optimize}$ denotes the algorithmic update strategy driven by the analyst's feedback, ensuring the system converges toward the global optimum while adhering to engineering constraints.

\subsection{\benchmark Construction}

To rigorously evaluate \framework in autonomous chemical process development, we present \benchmark, an engineering-verifiable benchmark repository. Unlike existing benchmarks limited to single-task accuracy, \benchmark targets the integrated closed-loop workflow of Knowledge Understanding, Structure Construction, and Parameter Optimization. This holistic benchmark mirrors the full lifecycle of chemical process design, ensuring that autonomous agents are assessed not merely on textual plausibility, but on rigorous engineering feasibility and validity.


\textbf{Knowledge Dimension.} To construct a robust cognitive foundation for autonomous agents, we established a dual-task evaluation framework comprising Knowledge Enhancement and Knowledge Extraction. This framework is grounded in a meticulously curated corpus of 70 core technical documents, which serves as the ground truth for industrial cognition.
The Knowledge Enhancement task is designed to rigorously assess the agent's reasoning fidelity in complex engineering scenarios. It necessitates the dynamic retrieval of specialized context to bridge the epistemic gap between general LLM capabilities and domain-specific engineering requirements. To quantify this capability, we developed a benchmark of 243 expert-level Question and Answer (Q\&A) pairs. Unlike generic retrieval tasks, these queries are strictly stratified into four critical dimensions of the chemical design lifecycle to ensure comprehensive coverage: (1) Process Route Selection: Evaluating strategic decision-making for optimal synthesis pathways; (2) Catalyst \& Condition Selection: Testing the precise identification of kinetic parameters and operating conditions; (3) Production Method \& Process Flow: Assessing the logical reconstruction of industrial-scale production sequences; and (4) Separation Method Selection: Verifying the correct application of thermodynamic separation technologies.
Complementing the reasoning assessment, the Knowledge Extraction task targets the structural formalization of unstructured technical text into machine-readable engineering data. To guide this transformation, we implemented a strict five-layer ontology that encompasses: (1) Process, (2) Flowsheet Structure, (3) Reaction Mechanism, (4) Substance, and (5) Constraints. This hierarchical schema enforces a rigorous standardization, requiring agents to precisely localize entities and resolve dependencies within the text. By mapping unstructured information to this constrained ontology, the framework effectively mitigates semantic drift and prevents the hallucination of non-existent chemical relationships, ensuring the high fidelity of the constructed knowledge base.


\textbf{Concept Dimension.} Building upon the knowledge foundation, this tier evaluates the agent's capability to manipulate process topologies through three integrated tasks: Parsing, Completion, and Generation.
First, the Parsing task targets the high-fidelity digitization of complex engineering schematics. We constructed a dataset of 113 sophisticated PFDs derived from the National College Student Chemical Design Competition, representing high-standard industrial practice. To establish a rigorous ground truth, expert annotators manually extracted the topological structure comprising equipment nodes and connectivity links from these images. This benchmark challenges agents to accurately perceive diagrammatic information and reconstruct the underlying graph structure from raw visual inputs.
Extending this structural perception, the Completion task evaluates the agent's deductive reasoning in incomplete information environments. Utilizing a subset of 69 high-quality samples from the parsing dataset, we randomly masked specific unit operations. The agent is tasked with identifying the masked equipment types and restoring the topological integrity based on the remaining context, thereby demonstrating deep engineering intuition and the ability to infer logical interconnects under uncertainty.
Finally, the Generation task assesses the constructive capability to translate unstructured process descriptions into executable PFD topologies. Chemical engineering experts curated representative industrial processes and synthesized them into unstructured textual summaries as input prompts. Adopting a Generation and Engineering Evaluation paradigm, the dataset comprises 30 distinct cases stratified into three complexity tiers to capture diverse scenarios: Simple (weakly coupled linear processes), Standard (reaction-separation series featuring local reflux loops), and Hard (strongly coupled systems characterized by multiple recycles and purge streams). This protocol enforces strict topological hard constraints to identify and penalize engineering failures, such as open loops or disconnected streams.


\textbf{Parameter Dimension.} The final tier of the benchmark, Chemical Process Parameter Optimization, represents the transition from static conceptual design to dynamic operational optimization. This module targets the evaluation of closed-loop, parameter-level decision-making within a fixed process topology. We constructed a rigorous evaluation environment comprising 20 high-fidelity scenarios, each anchored by an industrial-standard Aspen Plus backup file and a structured textual description of operational constraints. By integrating these thermodynamic simulators, the benchmark forces agents to interact with complex, non-convex physicochemical landscapes rather than simplified proxy models.
To ensure a comprehensive assessment of engineering capability, the optimization tasks are strictly stratified into three specific domains based on their objective functions. First, the Yield Optimization (8 cases) challenges the agent to maximize Product Flow Rate by navigating the reaction kinetic landscape, often requiring precise thermal control to suppress side reactions. Second, the Purity Optimization (9 cases) targets rigorous Molar Fraction standards, necessitating the fine-tuning of thermodynamic equilibrium in separation units to achieve pharmaceutical or polymer-grade specifications. Finally, the most challenging tier, Comprehensive Economic Optimization (3 cases), introduces a multi-objective trade-off. Here, the agent must minimize Annual Operating Cost, balancing the conflicting goals of maximizing product recovery while minimizing energy consumption (e.g., reboiler duty).
To traverse this complex solution space, the framework exposes a high-dimensional parameter set categorized into six critical engineering variables: Reactor Temperature, Reactor Length/Volume, and Residence Time for kinetic control; and Number of Plates, Reflux Ratio, and Feed Location for separation efficiency. This configuration rigorously tests the agent's ability to perform effective derivative-free optimization, relying on deep chemical intuition to iteratively adjust parameters and converge toward global optima within strict industrial boundaries.
A comprehensive statistical analysis detailing the distribution and composition of \benchmark is presented in Figure~\ref{fig:bench_statistics}.

\section*{Author contributions statement}
Conceptualization: Y.H.Y. and R.K.L.; Methodology: Y.H.Y., R.K.L., K.Z., and J.F.M.; Formal Analysis: Y.H.Y., R.K.L., and J.F.M.; Software: Y.H.Y., R.K.L., and J.F.M.; Investigation: Y.H.Y. and R.K.L.; Data Curation: Y.H.Y., R.K.L., J.F.M., J.Y.H., and Y.G.B.; Validation: Y.H.Y., R.K.L., K.Z., and J.F.M.; Visualization: Y.H.Y. and R.K.L.; Writing – Original Draft: Y.H.Y., R.K.L., and K.Z.; Writing – Review \& Editing: Y.H.Y., R.K.L., K.Z., Q.L., J.Y.H., J.B.Z., D.F.L., X.L., and E.H.C.; Supervision: K.Z.; Project Administration: K.Z.; Resources: K.Z. and E.H.C.; Funding Acquisition: K.Z. and E.H.C. All authors reviewed the manuscript.

\section*{Data availability}
The data supporting the findings of this study are available as follows. The datasets associated with the Knowledge and Parameter dimensions are publicly available in \url{https://github.com/kronos7777777/CeProAgents}. The datasets associated with the Concept dimension, sourced from the National College Student Chemical Design Competition, contain sensitive personal information. Due to privacy protection requirements, these data are not publicly deposited but are available from the corresponding author upon reasonable request.

\section*{Code availability}
All codes of proposed \framework are released in \url{https://github.com/kronos7777777/CeProAgents}.

\newpage
\bibliography{reference}



\newpage
\appendix 

\begin{center}
    \Large \textbf{Supplementary Information}
\end{center}

\setcounter{section}{0} 
\renewcommand{\thesection}{S\arabic{section}}

\setcounter{figure}{0}    
\renewcommand{\thefigure}{S\arabic{figure}}  

\setcounter{table}{0}
\renewcommand{\thetable}{S\arabic{table}}

\renewcommand{\thesection}{S\arabic{section}}
\renewcommand{\thefigure}{S\arabic{figure}}  
\renewcommand{\thetable}{S\arabic{table}}
\section{Extended Experimental Results}

\begin{figure}[h!]
\centering
\includegraphics[width=0.95\linewidth]{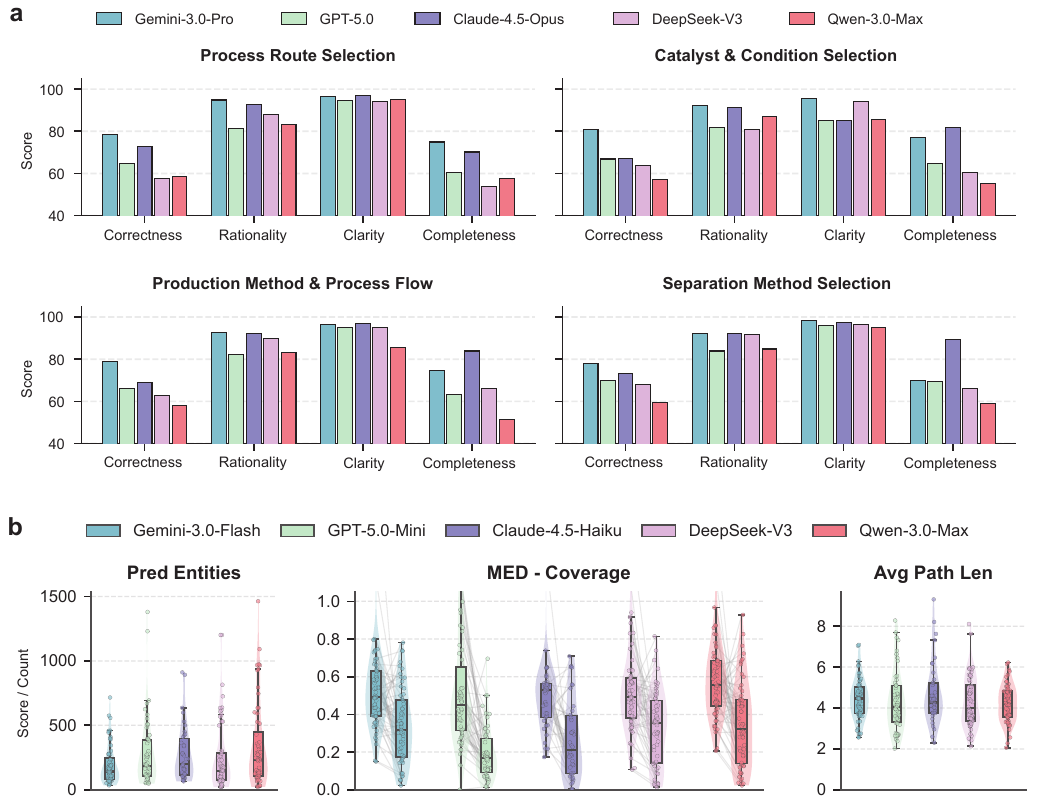}
\caption{\textbf{Extended experimental analysis of knowledge enhancement and extraction performance.}
\textbf{a,} Stratified evaluation of the Knowledge Augment task across four specific chemical engineering sub-domains: Process Route Selection, Catalyst \& Condition Selection, Production Method \& Process Flow, and Separation Method Selection, displaying performance scores for Correctness, Rationality, Clarity, and Completeness across five foundation models.
\textbf{b,} Statistical distribution of Knowledge Extraction metrics comparing different efficient backbone models, illustrating the density and variance for Predicted Entity Counts (left), Mapping-based Edge Distance (MED) Coverage (center), and Average Path Length of the constructed graphs (right).}
\label{fig:knowledge_Supplementary}
\end{figure}

\subsection{Extended Evaluation of the Knowledge Dimension}

In this section, we deepen our analysis of the \knowledgegroup by dissecting its performance across specific chemical engineering sub-domains and examining the statistical properties of the extracted knowledge graphs.

\textbf{Domain-Specific Knowledge Enhancement.} While aggregate metrics provide a high-level overview, the rigorous demands of chemical engineering require precise competencies in distinct sub-tasks. Figure~\ref{fig:knowledge_Supplementary}a stratifies the Knowledge Augment performance into four critical domains: Process Route Selection, Catalyst \& Condition Selection, Production Method, and Separation Method. Across all four categories, Gemini-3.0-Pro consistently establishes the performance ceiling, particularly in the Correctness dimension. Notably, in the highly sensitive ``Catalyst \& Condition Selection'' task, where hallucinated parameters can lead to experimental failure, the leading model achieves a Correctness score of 80.52, significantly outperforming the baseline average (e.g., DeepSeek-V3 at 63.60 and Qwen-3.0-Max at 57.32). A divergent trend is observed in Completeness. While models like Claude-4.5-Opus maintain strong Rationality (scoring 92.22 in Separation tasks), they exhibit a notable contraction in Completeness (dropping to 59.44), suggesting a tendency towards conservative, surface-level responses. In contrast, the Gemini-instantiated agent captures a broader spectrum of technical details (Completeness score of 70.00), ensuring that the retrieved context is sufficiently exhaustive for downstream decision-making.

\textbf{Statistical Characteristics of Structured Extraction.} Figure~\ref{fig:knowledge_Supplementary}b provides a granular view of the Knowledge Extraction process, comparing the distributional characteristics of the constructed graphs across different efficient backbones. The Predicted Entities distribution (left panel) highlights the recall capacity of the agents. Gemini-3.0-Flash exhibits a balanced distribution with a median entity count of approximately 133.5, indicating a robust ability to identify discrete chemical concepts without over-filtering. The MED - Coverage metric (center panel), which serves as a proxy for topological integrity, reveals that the model achieves a coverage score of 0.45, effectively mapping the extracted sub-graphs to the ground truth structure. Finally, the Average Path Length (right panel) measures the depth of the extracted semantic chains. The distribution converges around a path length of 4.11, suggesting that the system successfully constructs multi-hop causal chains rather than merely extracting isolated facts.
\begin{figure}[p]
\centering
\includegraphics[width=0.95\linewidth]{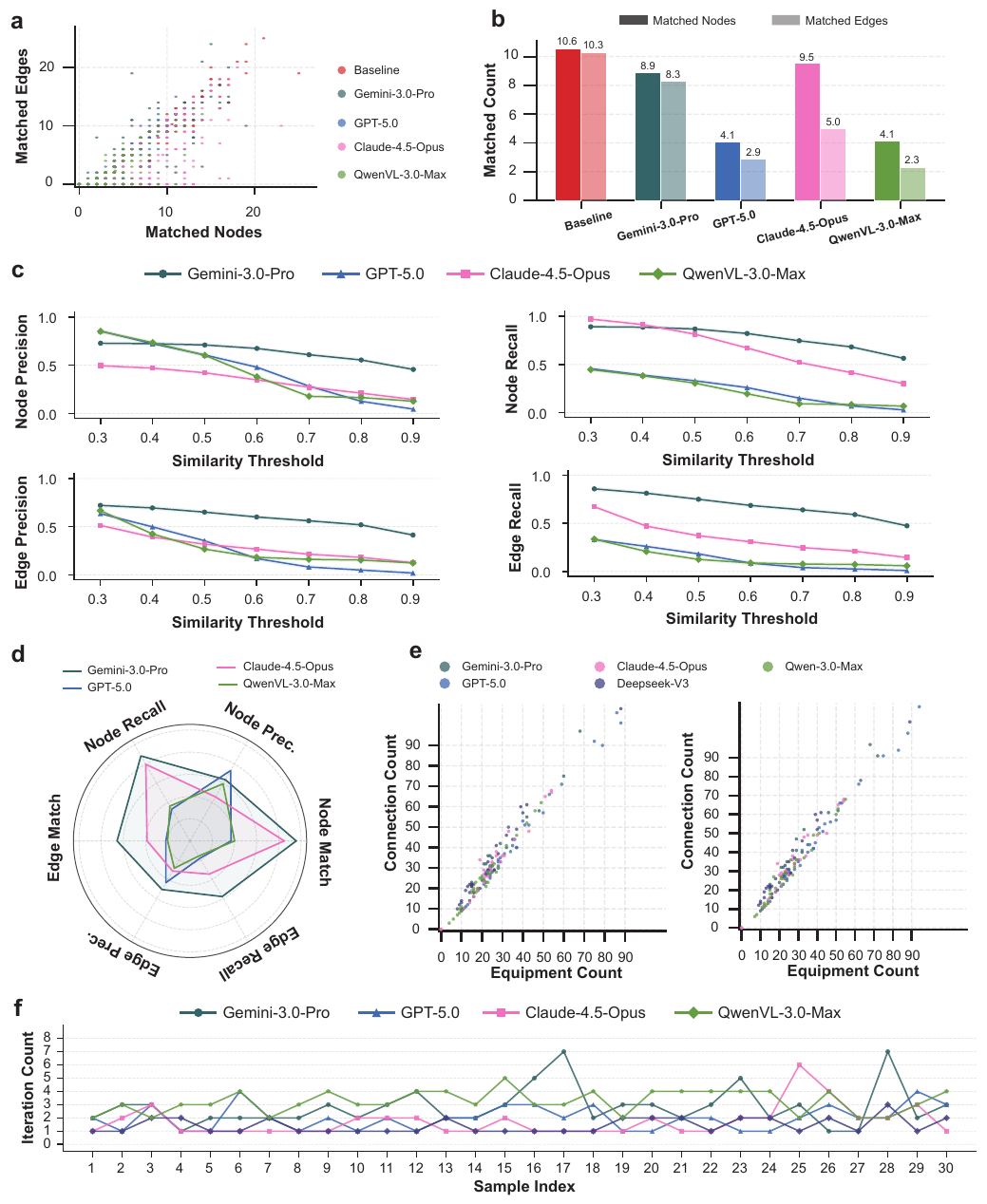}
\caption{\textbf{Extended experimental analysis of concept parsing, completion, and generative design.}
\textbf{a,} Distribution of matched edges versus matched nodes for concept parsing tasks across individual samples.
\textbf{b,} Average count of successfully matched nodes and edges for each model.
\textbf{c,} Variation of parsing precision and recall metrics as a function of the similarity threshold ($\tau$).
\textbf{d,} Parsing results for masked graph components in the topology completion task.
\textbf{e,} Comparative distribution of topological complexity (connections versus equipment) between the initial generation step and the final iteration.
\textbf{f,} Number of iterations required for the generative design process across distinct sample tasks.}
\label{fig:concept_Supplementary}
\end{figure}

\subsection{Extended Evaluation of the Concept Dimension}

In this section, we provide a granular analysis of the \conceptgroup's performance, systematically evaluating the robustness of topological parsing, the fidelity of structural extraction for completion tasks, and the evolutionary dynamics of the generative design process.


\textbf{Robustness of Topological Parsing.} To further validate the stability of our digitization framework under varying graph complexities, we analyzed the distribution of matched elements across the test set. As shown in Figure~\ref{fig:concept_Supplementary}a, the reasoning-strong backbones, particularly Gemini-3.0-Pro, exhibit a strong linear correlation between matched nodes and edges. This indicates a consistent parsing capability where the agent effectively captures the increasing density of connections as the number of equipment units grows, minimizing under-segmentation errors. In terms of absolute retrieval volume, Figure~\ref{fig:concept_Supplementary}b confirms that Gemini-3.0-Pro achieves the highest extraction density, identifying an average of 8.9 nodes and 8.3 edges per sample, closely approaching the ground truth baselines of 10.6 and 10.3, respectively.

We further scrutinized the system's sensitivity to matching strictness by varying the fuzzy similarity threshold ($\tau$) from 0.3 to 0.9 (Figure~\ref{fig:concept_Supplementary}c). While standard models experience a precipitous drop in performance as criteria tighten, the \framework architecture demonstrates remarkable stability. Notably, even at the stringent threshold of $\tau=0.9$, the Node Precision for the leading model remains above 0.45, and Edge Recall serves as a robust differentiator, maintaining a score of 0.47 compared to 0.06 for weaker backbones (e.g., QwenVL-3.0-Max). This confirms that our parsing workflow relies on semantic understanding rather than superficial string matching.

\textbf{Parsing Fidelity for Masked Topologies.} A critical prerequisite for the Concept Completion task is the accurate digitization of the incomplete graph structure surrounding the missing component. Figure~\ref{fig:concept_Supplementary}d quantifies the parsing metrics specifically for these masked diagrams. The radar chart reveals that the Gemini-instantiated agent establishes a comprehensive performance envelope, achieving balanced scores with Node Recall reaching 0.88 and Precision sustaining {0.64}. Crucially, the model maintains high Edge Recall (0.58) even in these compromised scenarios. This superior structural grounding ensures that the downstream Completion Agent receives a strictly accurate adjacency matrix, thereby maximizing the probability of correctly inferring the hidden unit operation based on its visible connectivity context.

\textbf{Evolutionary Dynamics of Generative Design.} To assess the impact of the adversarial correction mechanism on graph topology, Figure~\ref{fig:concept_Supplementary}e contrasts the distribution of connections versus equipment count between the initial generation step and the final optimized iteration. The initial designs (represented by the lighter distribution) often exhibit scattered topological properties with inconsistent connectivity densities (e.g., varying from 10 to 50 connections for 20 equipment units). However, through the multi-turn correction loop, the final topologies converge toward a structured linear alignment (slope = 1.2), reflecting the rigorous connectivity ratios observed in verified industrial PFDs.

The operational efficiency of this iterative process is quantified in Figure~\ref{fig:concept_Supplementary}f. The system demonstrates a decisive convergence behavior, typically resolving complex design constraints within 4 to 7 iterations. While some baselines exhibit oscillating iteration counts due to an inability to satisfy engineering checks, the Gemini-3.0-Pro backbone minimizes computational overhead by reaching valid terminal states with an average of 4.5 steps, effectively balancing exploration speed with engineering rigor.

\begin{figure}[p]
\centering
\includegraphics[width=0.95\linewidth]{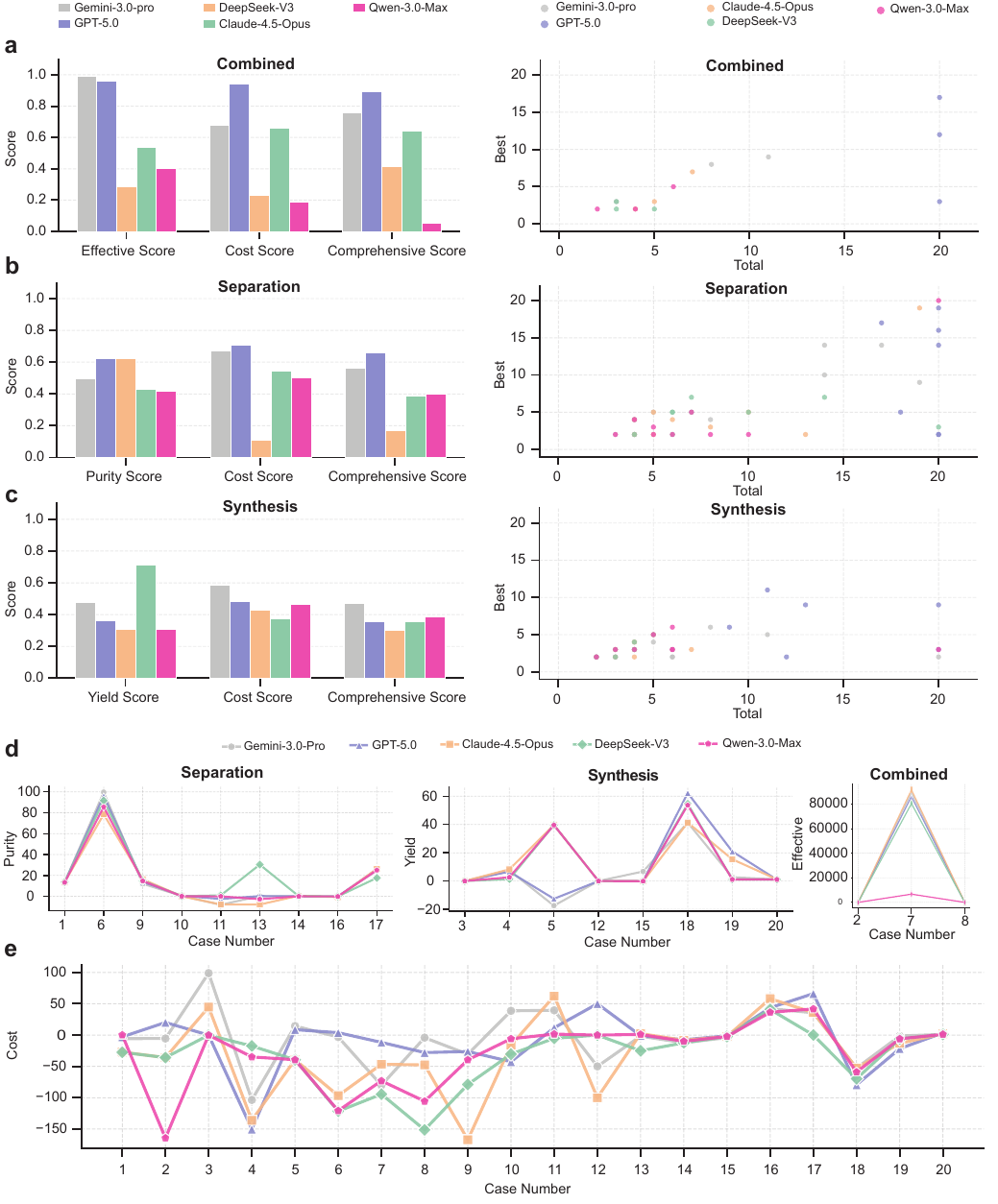}
\caption{\textbf{Extended quantitative statistics for parameter optimization across distinct chemical process types.}
\textbf{a-c,} Stratified analysis of optimization performance and search efficiency: \textbf{a,} Combined processes; \textbf{b,} Separation processes; and \textbf{c,} Synthesis processes. In each panel, the left bar charts display the normalized performance scores across five key metrics, while the right scatter plots correlate the iteration step where the global optimum was identified versus the total iterations performed.
\textbf{d,} Line graphs tracking the absolute performance trajectories for primary objectives, including Purity (Separation), Yield (Synthesis), and Effective Score (Combined), across individual benchmark cases.
\textbf{e,} Longitudinal analysis of economic optimization displaying the raw Operating Cost values recorded across 20 distinct test cases for different model backbones.}
\label{fig:parameter_Supplementary}
\end{figure}

\subsection{Extended Evaluation of the Parameter Dimension}

This section provides a comprehensive assessment of the \parametergroup's ability to autonomously optimize industrial process parameters. We categorize the evaluation into distinct unit operation types, Combined, Separation, and Synthesis, to isolate specific reasoning capabilities.

\textbf{Stratified Performance Analysis.} Figures~\ref{fig:parameter_Supplementary}a-c (left panels) illustrate the comparative performance across the three process categories. In Synthesis tasks (Figure~\ref{fig:parameter_Supplementary}c), the performance gap is relatively narrow, with most backbones achieving Comprehensive scores clustered around 0.35 to 0.45. This suggests that maximizing reaction kinetics is a task that benefits broadly from general reasoning capabilities. However, a significant divergence emerges in the more thermodynamically complex Separation (Figure~\ref{fig:parameter_Supplementary}b) and Combined (Figure~\ref{fig:parameter_Supplementary}a) tasks. Here, reasoning-strong backbones like Gemini-3.0-Pro and GPT-5.0 demonstrate superior proficiency, achieving Effective Scores of approximately 0.98 and 0.95 respectively, whereas weaker models exhibit a notable contraction, dropping to scores below 0.55.

\textbf{Search Efficiency and Convergence.} The scatter plots in Figures~\ref{fig:parameter_Supplementary}a-c (right panels) provide insight into the agents' decision-making decisiveness by correlating the iteration step where the global optimum was found (Best) against the total iteration budget consumed (Total). The data reveals that Gemini-3.0-Pro typically locates the optimal solution within the first 2 to 4 iterations and terminates decisively. In contrast, other models exhibit suboptimal behaviors such as premature termination (e.g., stopping at 5 iterations before convergence) or inefficient over-exploration (e.g., GPT-5.0 consistently exhausting the budget of 20 iterations), as evidenced by the scatter distribution relative to the diagonal ideal.

\textbf{Stability and Economic Optimization.} To evaluate robustness under diverse boundary conditions, Figures~\ref{fig:parameter_Supplementary}d and~\ref{fig:parameter_Supplementary}e track the specific performance trajectories across individual test cases. Figure~\ref{fig:parameter_Supplementary}d displays the raw objective values for Purity, Yield, and Effective. While baseline models suffer from significant volatility, evident in the sharp dips in performance for Case 5 (Synthesis) and Case 10 (Separation), the leading models maintain a stable high-performance envelope. Furthermore, Figure~\ref{fig:parameter_Supplementary}e analyzes the economic dimension by plotting the absolute Operating Cost profiles. The significant variability observed in cost minimization, particularly in complex cases like Case 4 (ranging from -150 to 100), underscores the non-convex nature of the optimization landscape. The leading agents successfully navigate these trade-offs, consistently identifying operating regions that minimize real-world costs while satisfying engineering constraints.

\section{Statistical Analysis of \benchmark}

To ensure the generalizability and industrial relevance of \framework, we constructed \benchmark, a multi-modal repository spanning the full lifecycle of chemical engineering. This section provides a comprehensive statistical profiling of the benchmark's three core dimensions, demonstrating its diversity in data scale, topological complexity, and optimization difficulty.

\begin{figure}[p]
\centering
\includegraphics[width=1\linewidth]{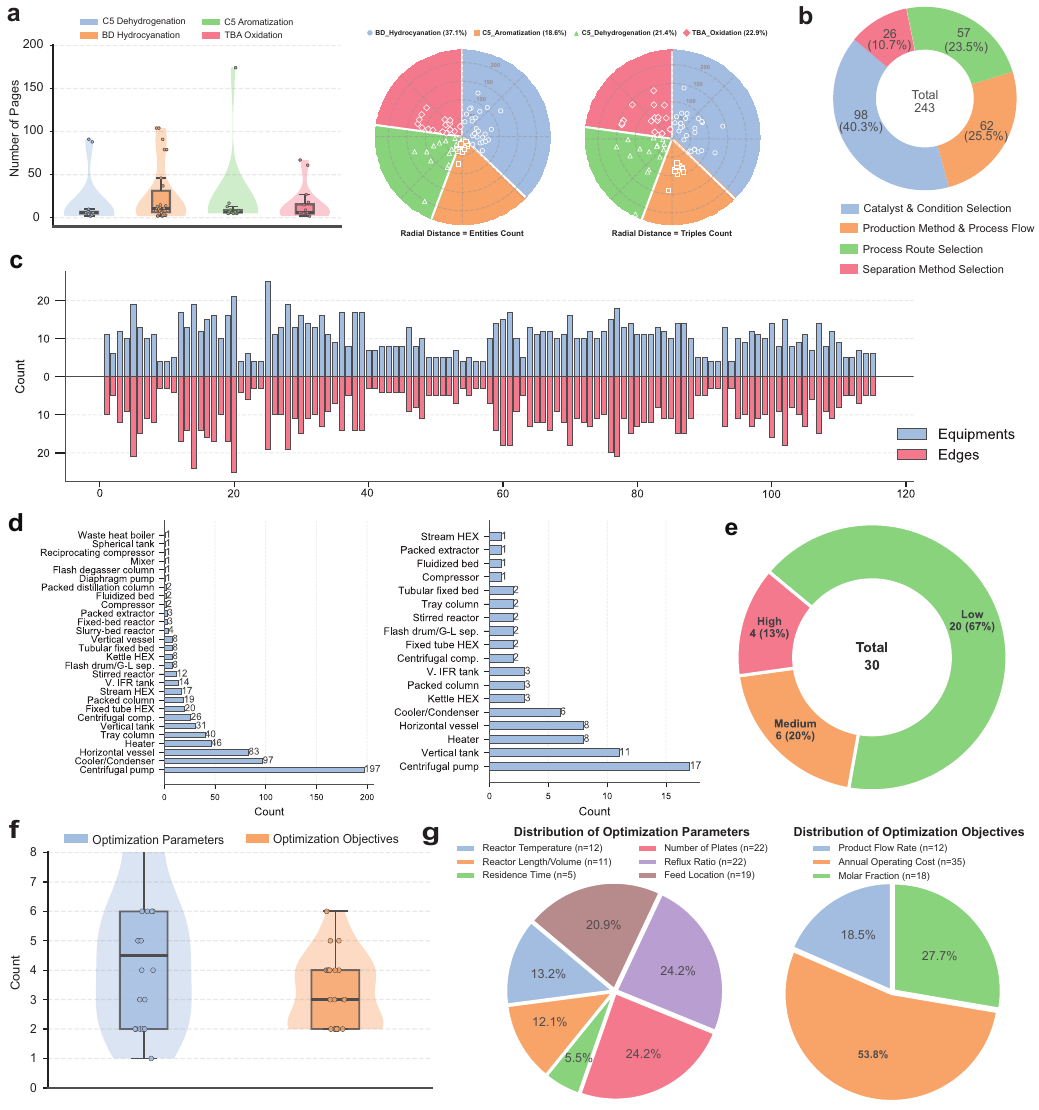}
\caption{\textbf{Comprehensive statistical profiling of the \benchmark.} 
    \textbf{a,} Statistical characteristics of the Knowledge dimension corpus across industrial sub-domains, illustrating the distribution of document length, the density of entities and triples, and their correlation, validating semantic consistency.
    \textbf{b,} Composition of the Q\&A dataset, detailing the distribution of task types and difficulty scores, highlighting the prevalence of high-complexity reasoning tasks in specific categories.
    \textbf{c,} Topological complexity of the PFD dataset, quantified by the count of equipment nodes and connection edges per sample.
    \textbf{d,} Frequency distribution of unit operations, contrasting the overall equipment distribution with the subset targeted for topology completion tasks, reflecting a focus on critical process units.
    \textbf{e,} Difficulty stratification of the generative text-to-design scenarios.
    \textbf{f,} Dimensionality analysis of the Simulation scenarios, showing the distribution of optimization parameters and objectives.
    \textbf{g,} Detailed breakdown of optimization parameters and objectives, emphasizing the benchmark's focus on separation efficiency and economic optimization.}
\label{fig:bench_statistics}
\end{figure}

\textbf{Knowledge Dimension Characterization.} The Knowledge dimension is anchored by a curated corpus spanning four industrial domains. As shown in Figure~\ref{fig:bench_statistics}a, the corpus exhibits significant structural heterogeneity with technical reports extending up to 200 pages, necessitating robust long-context processing capabilities. Despite this diversity, the radial distribution of entity and triple counts illustrates balanced complexity across all domains. The associated Q\&A dataset (Figure~\ref{fig:bench_statistics}b) is dominated by ``Catalyst \& Condition Selection'' tasks, while reasoning-intensive categories like ``Separation Method Selection'' are predominantly high-difficulty, serving as a rigorous stress test for agentic reasoning.

\textbf{Concept Dimension Characterization.} The Concept dimension is supported by high-fidelity PFDs derived from industrial competitions. Statistical analysis of the graph topology (Figure~\ref{fig:bench_statistics}c) demonstrates significant variance in problem size, with equipment nodes and connection edges analyzed across distinct samples. The equipment ontology distribution (Figure~\ref{fig:bench_statistics}d) aligns with industrial frequency, dominated by transport and thermal regulation units (``Centrifugal pumps'', ``Coolers/Condensers''). However, for topology completion tasks, we specifically target critical unit operations such as ``Kettle HEX'' and ``Fixed-bed reactors'' to probe deep engineering intuition rather than simple pattern matching. For generative design, the text-to-flowsheet scenarios are tiered by topological complexity (Figure~\ref{fig:bench_statistics}e). Notably, a significant portion of tasks are classified as high-difficulty, requiring the synthesis of multi-loop systems with coupled recycle streams and stringent connectivity constraints.

\textbf{Parameter Dimension Characterization.} The Parameter dimension comprises high-fidelity simulation scenarios integrated with Aspen Plus. As shown in Figure~\ref{fig:bench_statistics}f, the problem space is high-dimensional; each scenario typically mandates the simultaneous optimization of multiple parameters to satisfy conflicting optimization objectives. The distribution of decision variables (Figure~\ref{fig:bench_statistics}g) is heavily weighted towards separation logic, with ``Number of Plates'' and ``Reflux Ratio'' dominating, reflecting the non-linear thermodynamic challenges of azeotropic and extractive distillation. Correspondingly, ``Annual Operating Cost'' serves as the primary optimization objective, ensuring that the optimization landscape enforces a rigorous trade-off between engineering performance (Yield/Purity) and economic viability.

\section{Definition of Evaluation Metrics}
To rigorously assess the performance of the \framework system, we established a multi-dimensional evaluation framework. Selected metrics across the Knowledge, Concept, and Parameter dimensions are defined as follows:

\begin{itemize}
    \item \textbf{Knowledge Extraction Metrics.} We evaluate entity extraction using Accuracy ($A$), Recall ($R$), and F1-score ($F1$). Let $N_{match}$ be the number of matched entities, $N_{pred}$ the total predicted, and $N_{gt}$ the ground truth:
    \begin{equation}
        A = \frac{N_{match}}{N_{pred}}, \quad R = \frac{N_{match}}{N_{gt}}, \quad F1 = \frac{2 \cdot A \cdot R}{A + R}.
    \end{equation}
    For graph topology, we calculate Mapping-based Edge Connectivity ($MEC$) and Edge Distance ($MED$). Here, $E_{gt}$ is the set of ground-truth edges, $E_{conn}$ is the subset of connected edges, and $d(u,v)$ is the shortest path distance in the predicted graph normalized by the average distance $\bar{d}$:
    \begin{equation}
        MEC = \frac{1}{|E_{gt}|} \sum_{(u,v) \in E_{gt}} \delta(u, v), \quad MED = \frac{1}{|E_{conn}|} \sum_{(u,v) \in E_{conn}} \frac{d(M(u), M(v))}{\bar{d}}.
    \end{equation}

    \item \textbf{Knowledge Enhancement Metrics.} We evaluate reasoning quality independently across four dimensions ($D$): Correctness, Rationality, Clarity, and Completeness. For a specific query $Q$, answer $A$, and reference $R$, the score $S_d$ for dimension $d \in D$ is assigned by the LLM judge:
    \begin{equation}
        S_d = \text{Judge}(Q, A, R \mid d).
    \end{equation}

    \item \textbf{Concept Parsing Metrics.} We measure the parsing fidelity for Equipments ($Eq$) and Connections ($Cn$) using Accuracy ($A$) and Recall ($R$). A match $m(x, y)$ is binary (1 or 0) based on a similarity threshold:
    \begin{equation}
        A_{Eq} = \frac{\sum m(p, g)}{|Eq_{pred}|}, \quad R_{Eq} = \frac{\sum m(p, g)}{|Eq_{gt}|}, \quad A_{Cn} = \frac{\sum m(c_{pred}, c_{gt})}{|Cn_{pred}|}, \quad R_{Cn} = \frac{\sum m(c_{pred}, c_{gt})}{|Cn_{gt}|}.
    \end{equation}

    \item \textbf{Concept Completion Metrics.} We evaluate the prediction of masked components using Top-K Accuracy ($A@K$). Let $y$ be the ground truth and $\hat{Y}_{K}$ be the top-$K$ predicted candidates:
    \begin{equation}
        A@K = \delta(y \in \hat{Y}_{K}).
    \end{equation}

    \item \textbf{Concept Design Metrics.} We evaluate generative quality using Valid Rate ($R_{valid}$) for topological legality and Correct Rate ($R_{correct}$) for engineering compliance. Let $N_{total}$ be the total generated designs:
    \begin{equation}
        R_{valid} = \frac{N_{valid}}{N_{total}}, \quad R_{correct} = \frac{N_{correct}}{N_{total}}.
    \end{equation}

    \item \textbf{Parameter Optimization Metrics.} We measure performance using the improvement ratios ($r$) of Yield ($Y$), Purity ($P$), and Cost ($C$) relative to their initial states ($init$). We further define Effective ($r_{eff}$) and Overall ($r_{overall}$) metrics to capture trade-offs:
    \begin{equation}
        r_Y = \frac{Y_{opt}}{Y_{init}}, \quad r_P = \frac{P_{opt}}{P_{init}}, \quad r_C = \frac{C_{opt}}{C_{init}}
        , \quad r_{eff} = r_Y \cdot r_P, \quad r_{overall} = \frac{r_{eff}}{r_C}.
    \end{equation}
    The final reported scores ($S$) correspond to the normalized values of these ratios across the evaluated models.
\end{itemize}

\newpage
\section{Detailed of Cases}
\begin{figure}[h!]
\centering
\includegraphics[width=\linewidth]{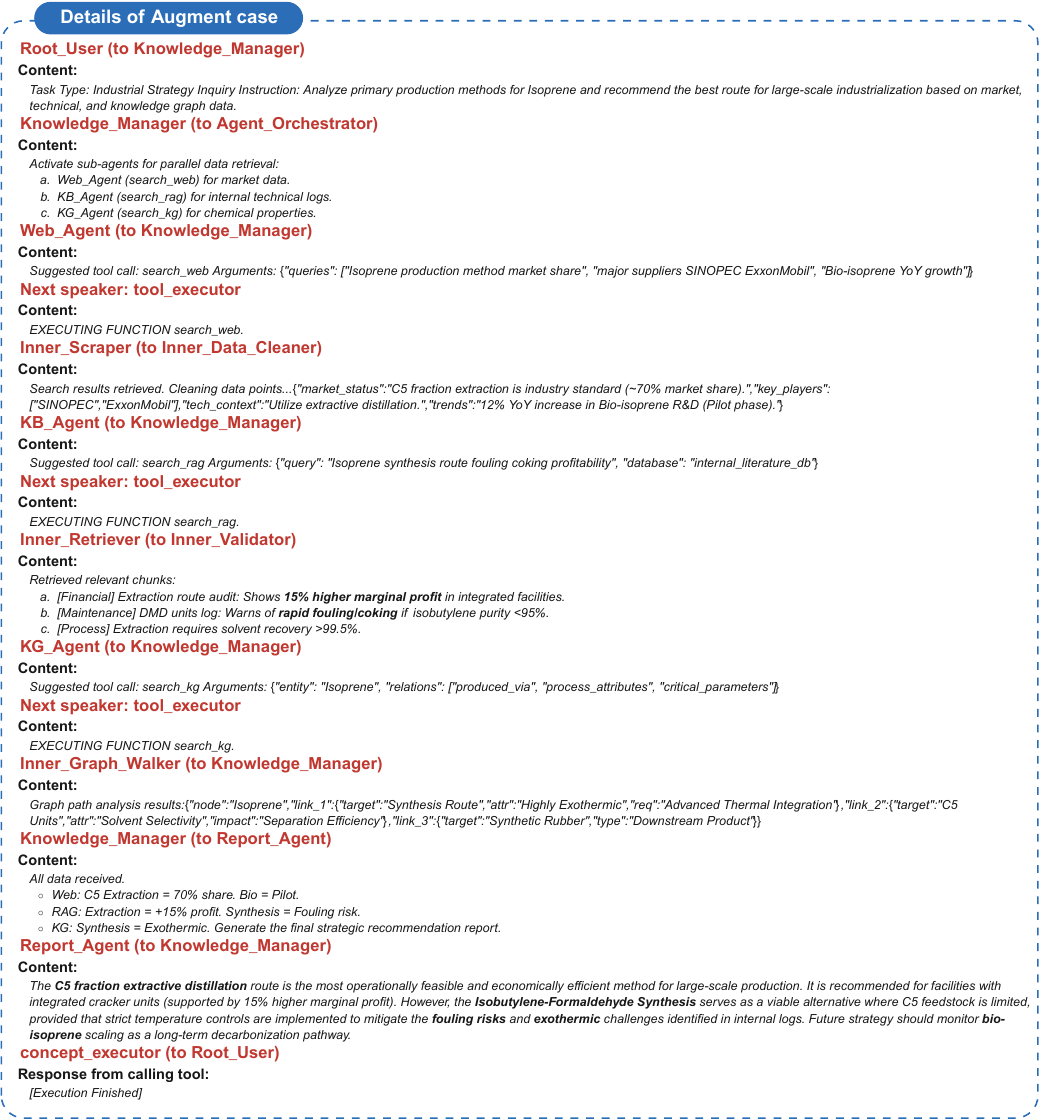}
\end{figure}

\begin{figure}[t!]
\centering
\includegraphics[width=\linewidth]{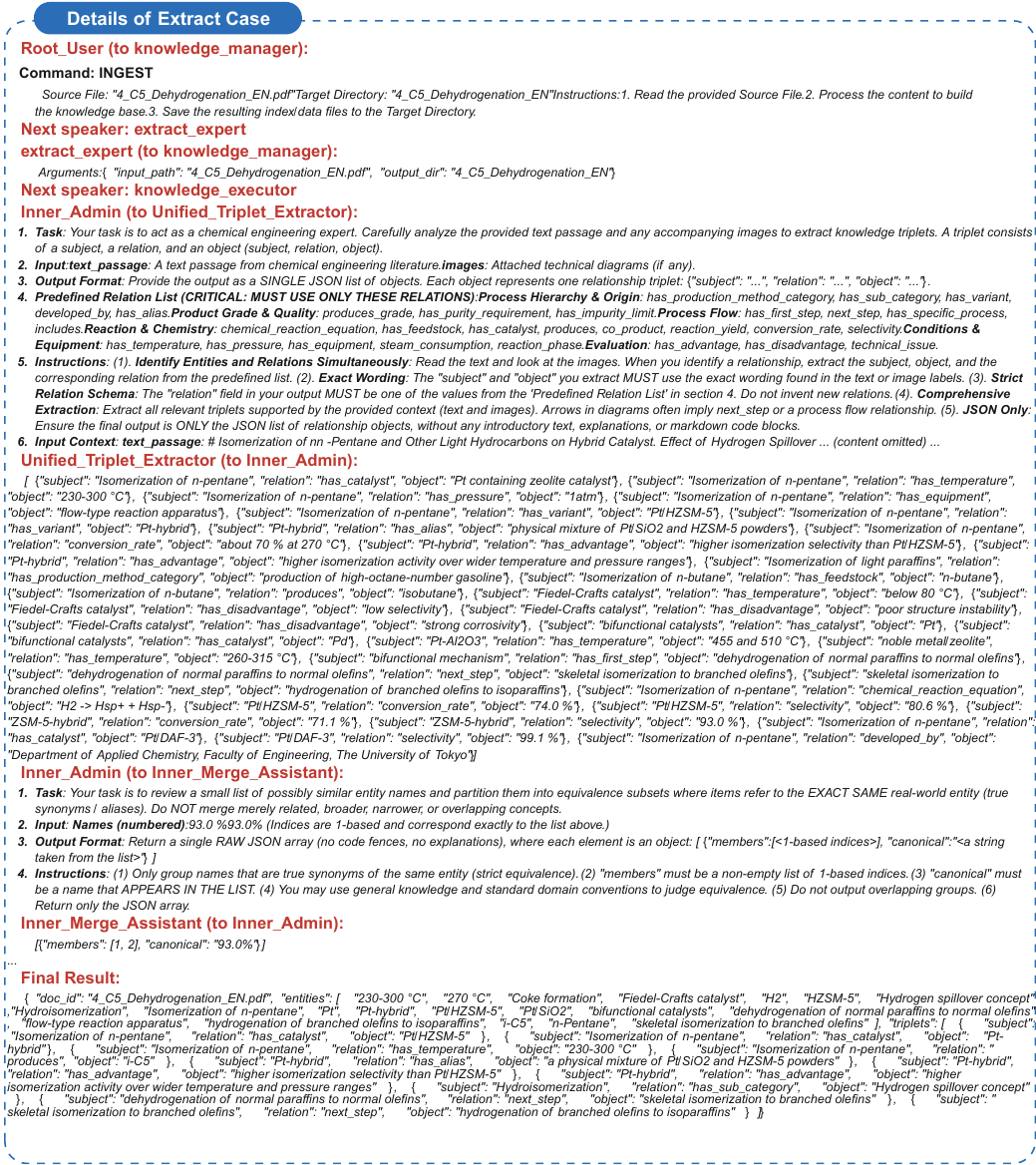}
\end{figure}

\begin{figure}[t!]
\centering
\includegraphics[width=\linewidth]{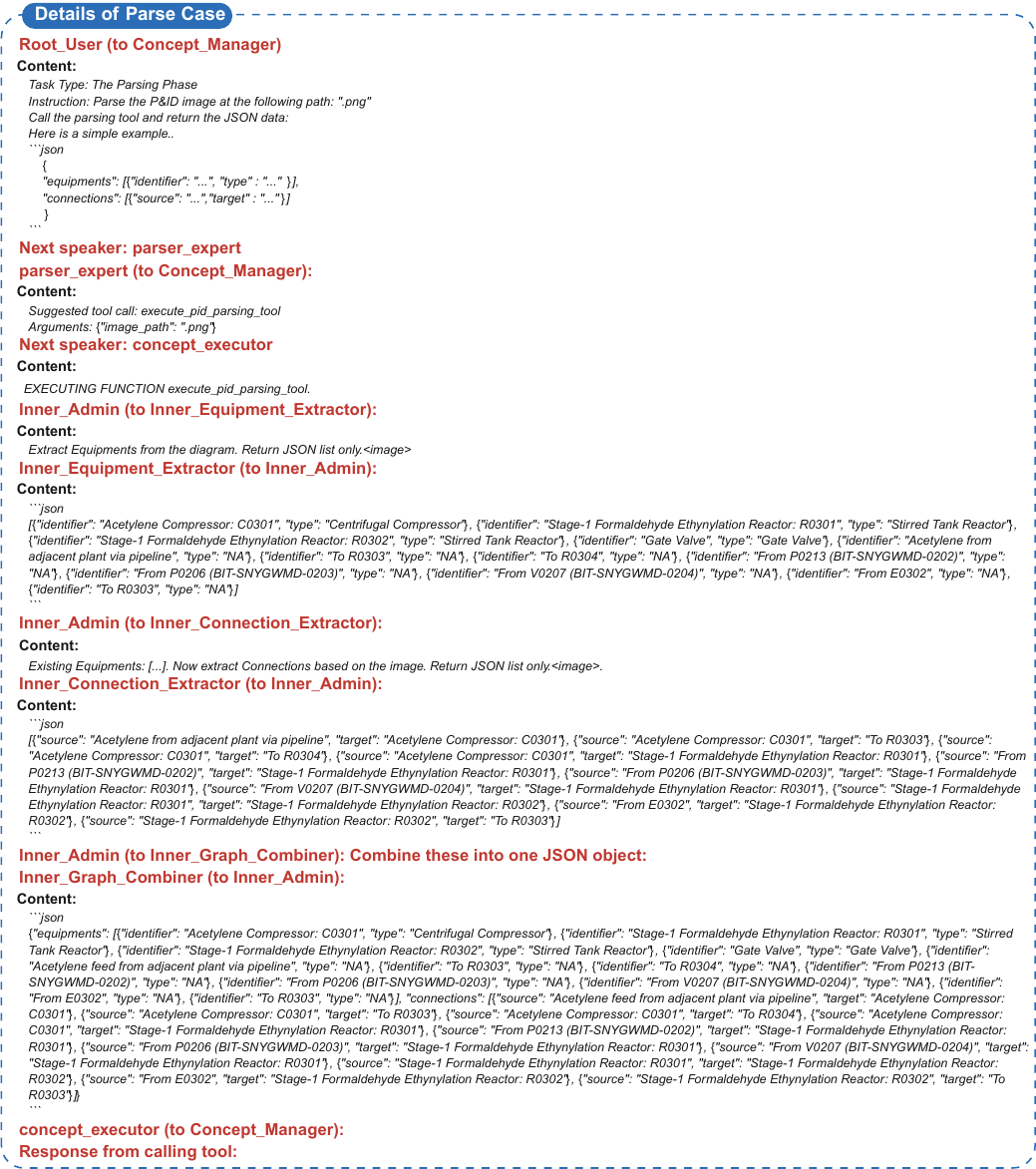}
\end{figure}

\begin{figure}[t!]
\centering
\vspace{-1.2cm} 
\includegraphics[width=\linewidth]{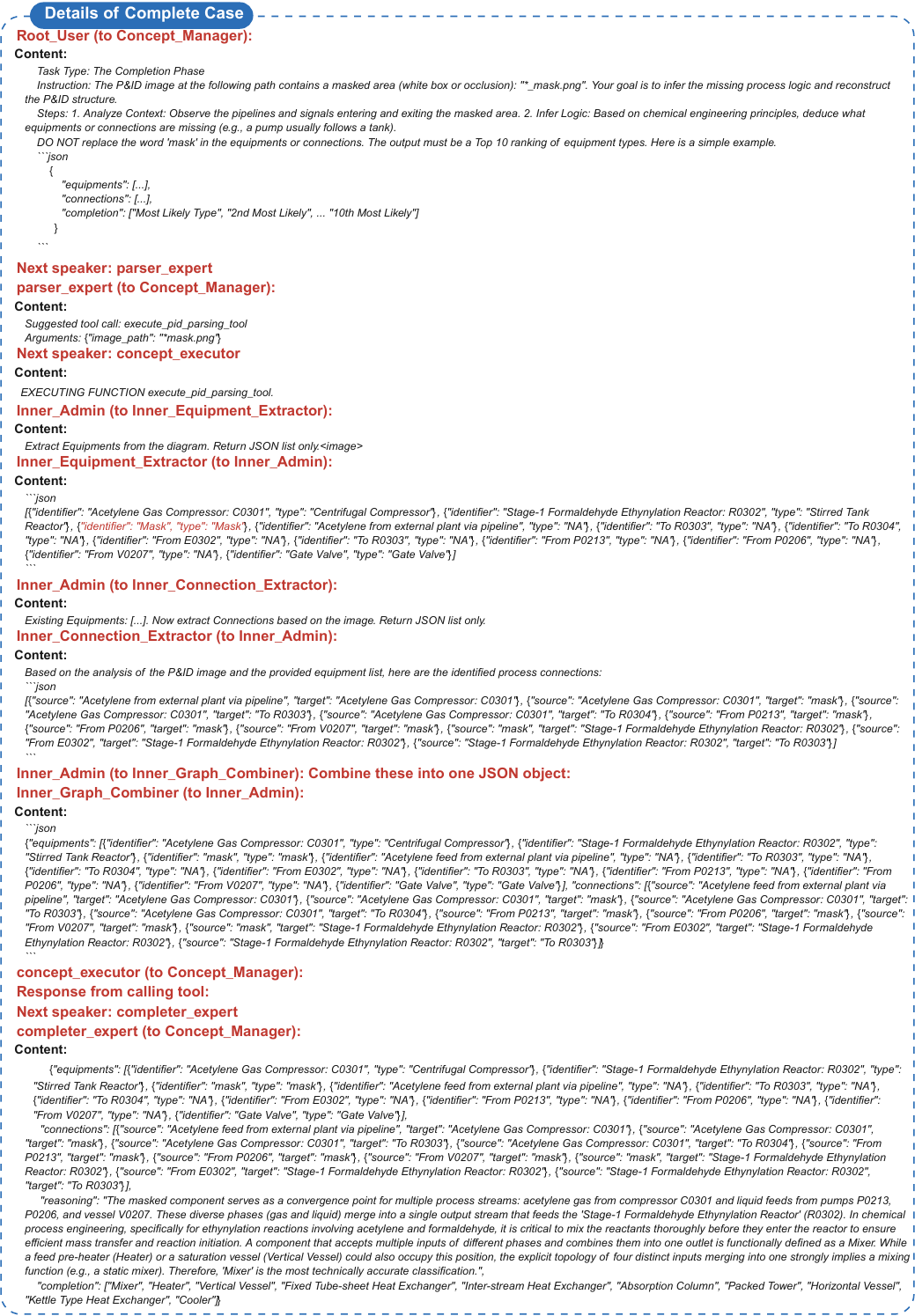}
\end{figure}

\begin{figure}[t!]
\centering
\vspace{-1.2cm} 
\includegraphics[width=\linewidth]{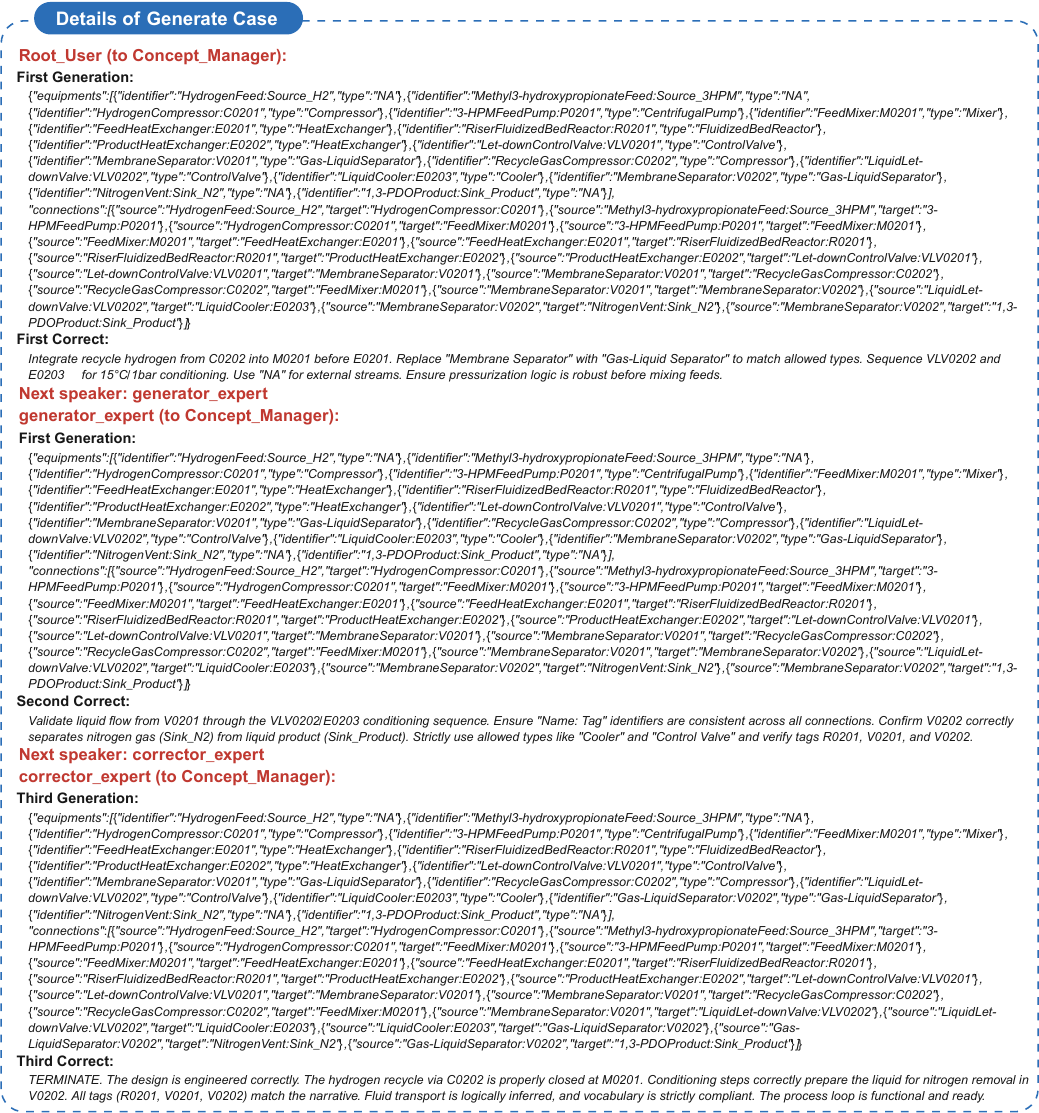}
\end{figure}

\begin{figure}[t!]
\centering
\vspace{-3cm} 
\includegraphics[width=\linewidth]{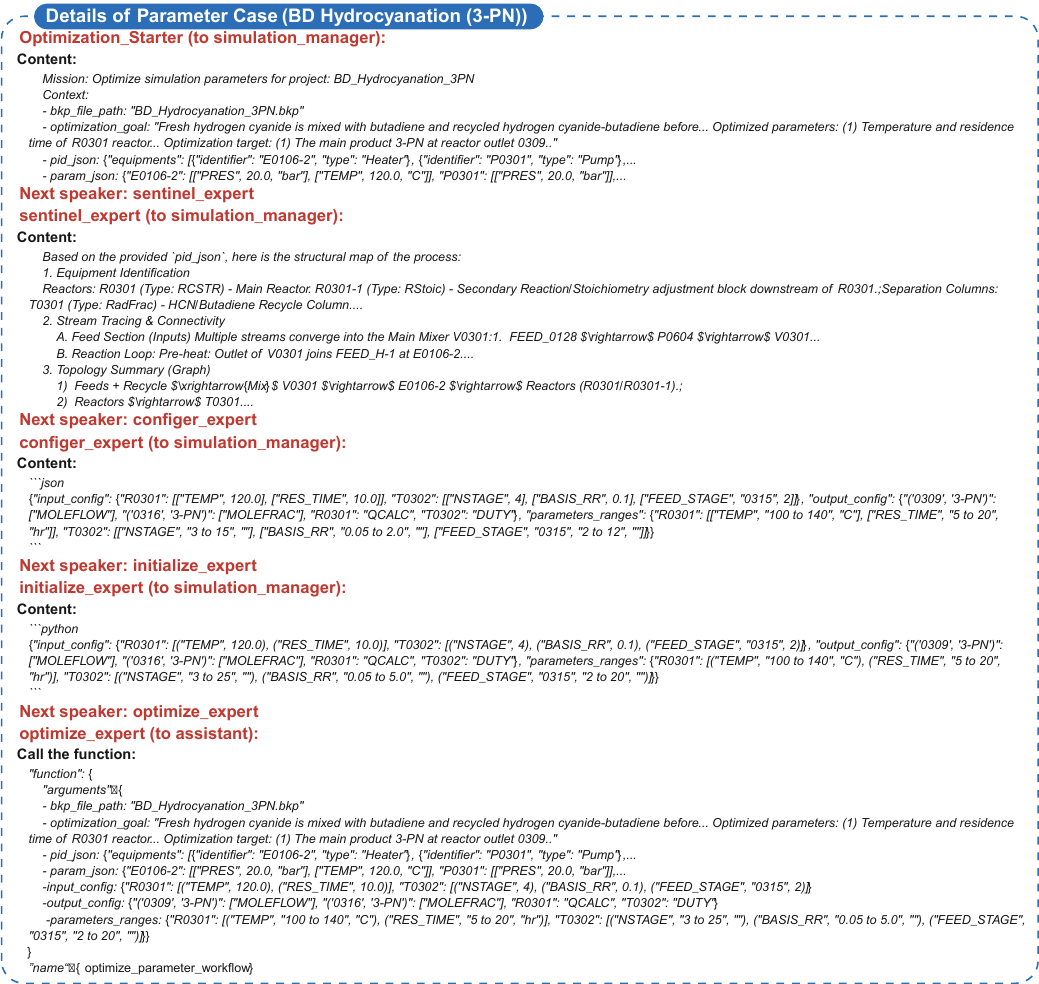}
\end{figure}

\begin{figure}[t!]
\centering
\includegraphics[width=\linewidth]{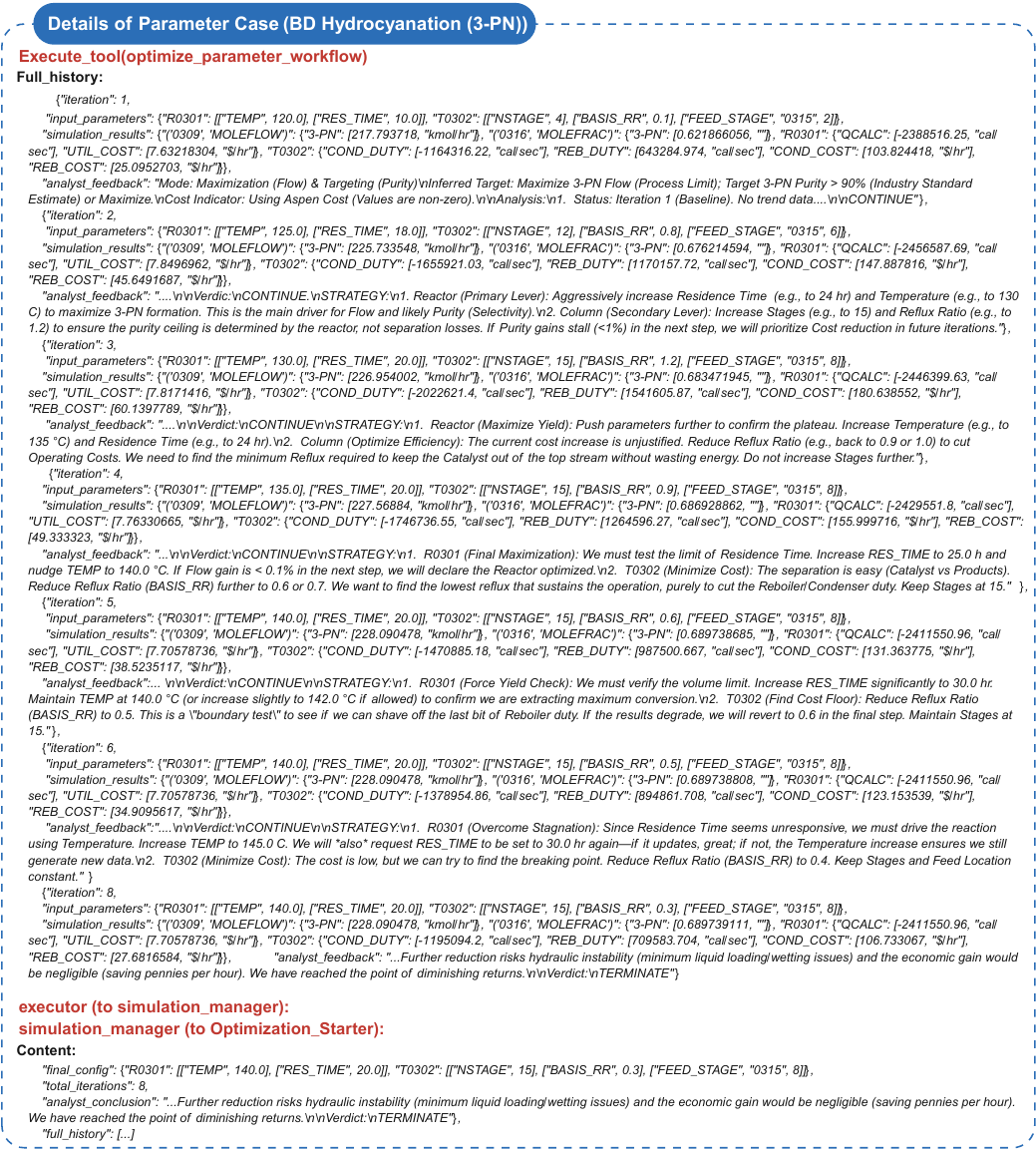}
\end{figure}

\begin{figure}[t!]
\centering
\vspace{-1cm} 
\includegraphics[width=\linewidth]{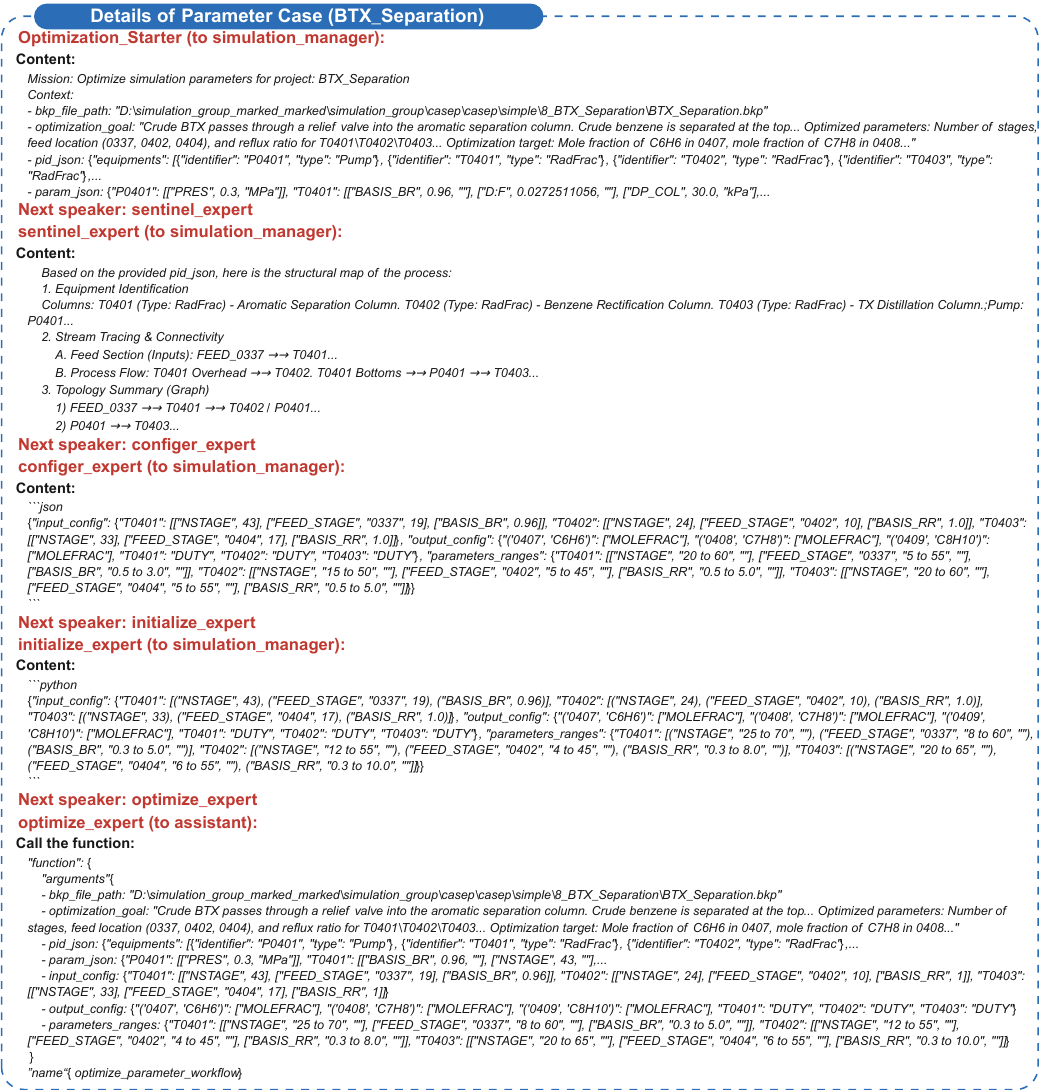}
\end{figure}

\begin{figure}[t!]
\centering
\vspace{-4cm} 
\includegraphics[width=\linewidth]{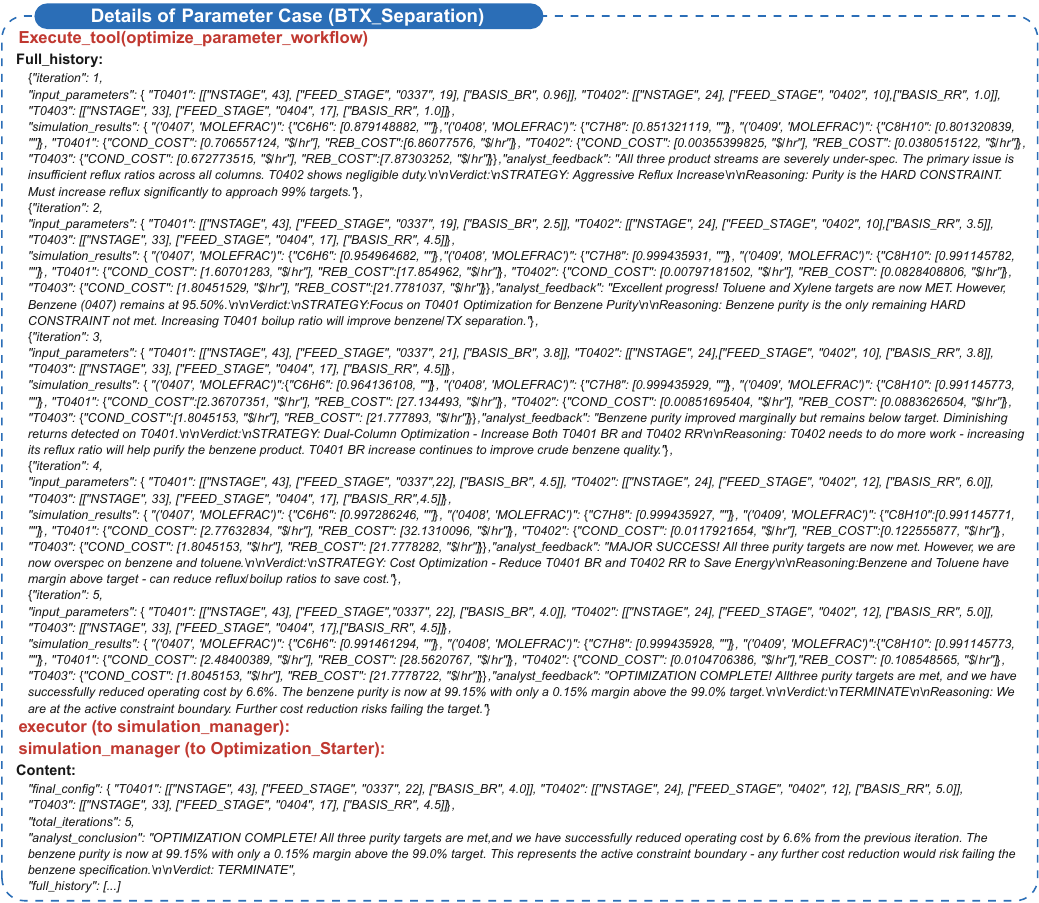}
\end{figure}

\begin{figure}[t]
\centering
\includegraphics[width=\linewidth]{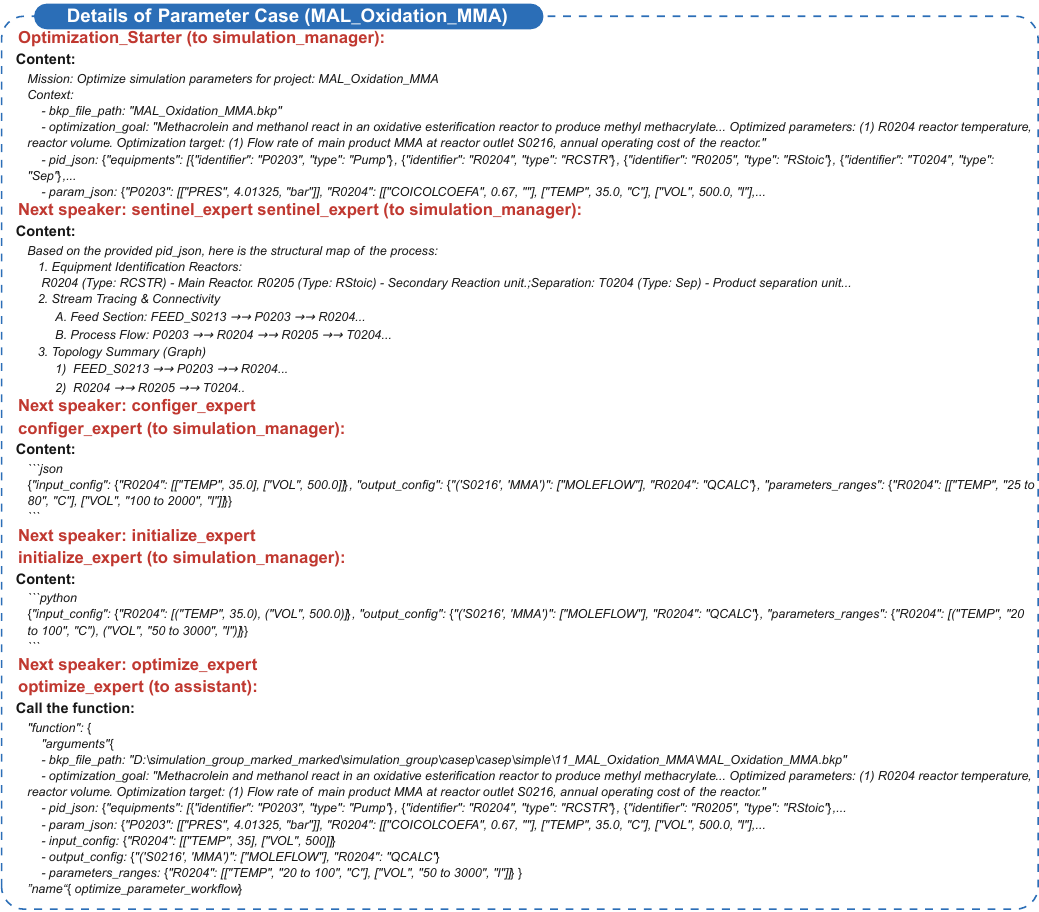}
\end{figure}

\begin{figure}[t!]
\centering
\includegraphics[width=\linewidth]{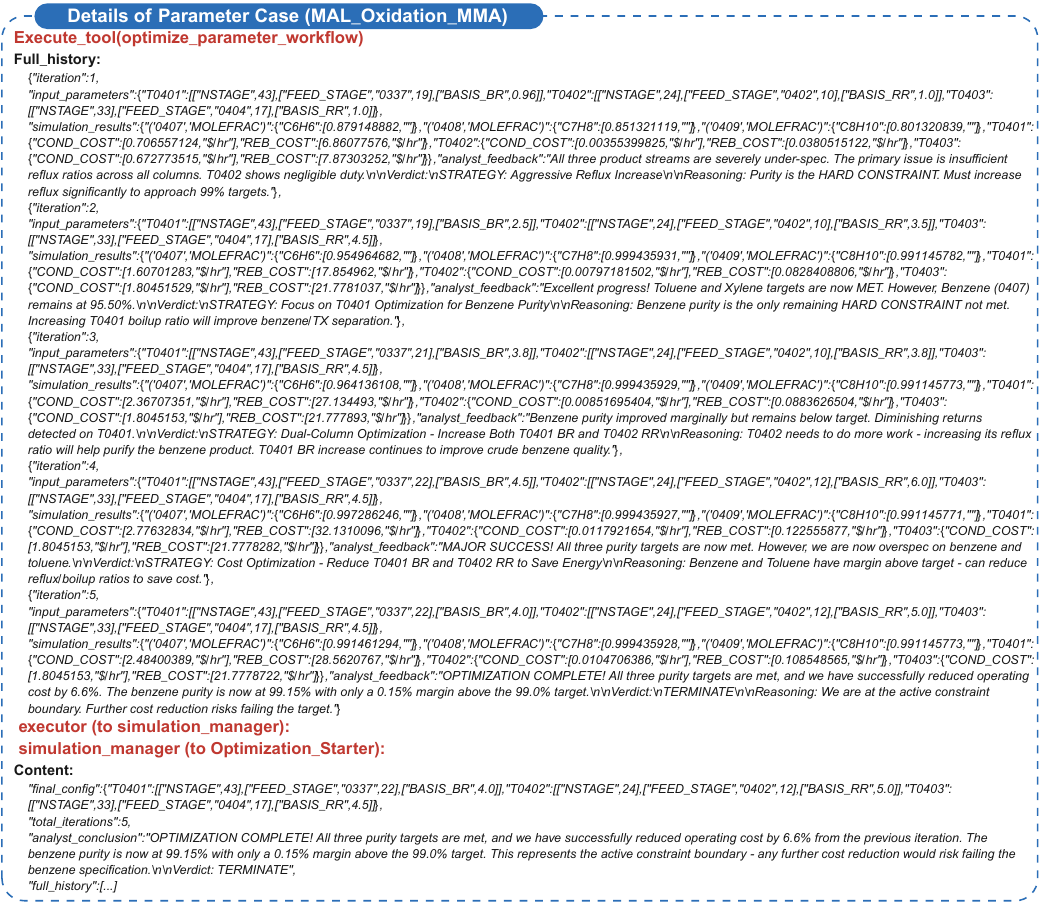}
\end{figure}

\newpage
\section{Methodological Details of \framework}

\subsection{Details of Agents Designing}
In this section, we present the detailed design of agents in \framework.

\textbf{\knowledgegroup.}
To provide the system with robust epistemic foundations and mitigate the inherent hallucinatory tendencies of generative models, we instantiate a specialized \knowledgegroup $S_{knowledge}$. This cohort serves as the dynamic memory and information retrieval engine of \framework, bridging the gap between the static parameters of the base LLMs and the rapidly evolving frontier of scientific literature. $S_{knowledge}$ is architecturally partitioned into a deliberation-based Knowledge Augment ChatGroup $G_{knowledge}$ for multi-source evidence synthesis and a deterministic Knowledge Update Workflow $F_{knowledge}$ for continuous data enrichment.
$G_{knowledge}$ is configured to perform multi-modal information retrieval through four specialized agents. Each agent is equipped with a distinct tool to query heterogeneous data sources: 
\begin{itemize}
    \item \textbf{Web Search Agent} ($a_{web}$) utilizes the DuckDuckGo engine (DDGS) to capture real-time developments and grey literature;
        \begin{figure}[ht]
            \centering
            \vspace{-0.5cm} 
            \includegraphics[width=0.9\linewidth]{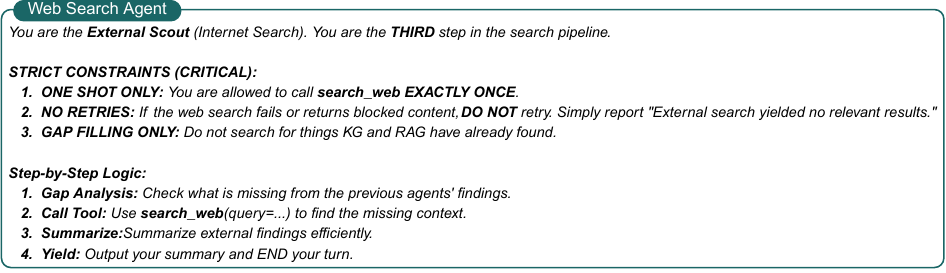}\
            \caption*{}
        \end{figure}
        \vspace{-1.2cm} 
    \item \textbf{Knowledge Graph Agent} ($a_{kg}$) interfaces with a Neo4j graph database to retrieve structured, ontological relationships between chemical entities;
        \begin{figure}[ht]
            \centering
            \includegraphics[width=0.9\linewidth]{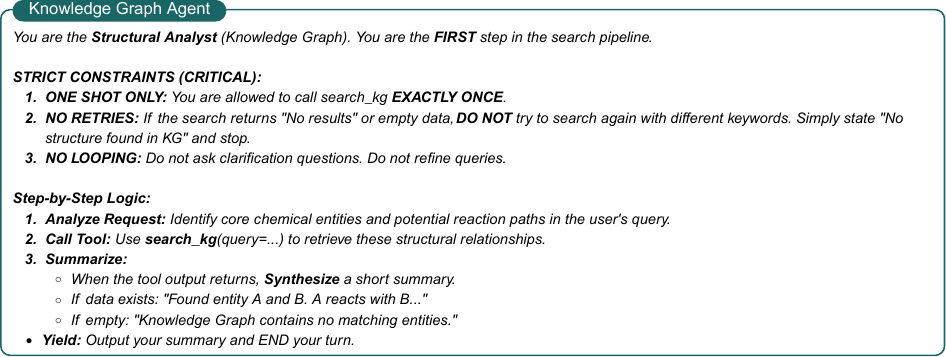}\
            \caption*{}
        \end{figure}
    \vspace{-1cm} 
    \item \textbf{Knowledge Base Agent} ($a_{kb}$) employs a Retrieval-Augmented Generation (RAG) pipeline via ChromaDB to access domain-specific internal corpora. To ensure the veracity of the retrieved information;
        \begin{figure}[ht]
            \centering
            \includegraphics[width=0.9\linewidth]{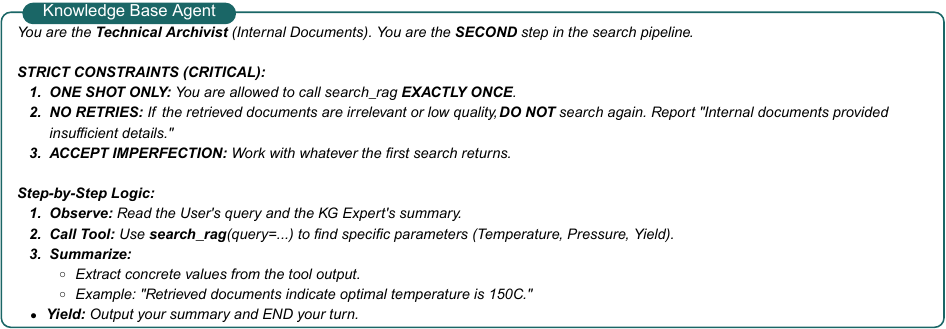}\
            \caption*{}
        \end{figure}
    \item \textbf{Report Agent} ($a_{re}$) acts as a cognitive synthesizer to integrate these heterogeneous streams and resolve conflicting data points.
        \begin{figure}[ht]
            \centering
            \includegraphics[width=0.9\linewidth]{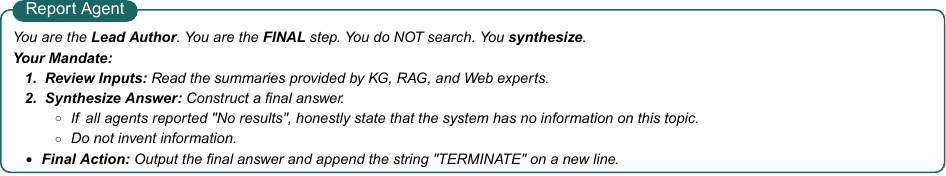}\
            \caption*{}
        \end{figure}
    \item \textbf{Knowledge Extraction Agent}($a_{ke}$) operates as the core of the ingestion pipeline, transforming unstructured chemical engineering text into structured data. 
\end{itemize}
This collaborative structure enables $G_{knowledge}$ to transform fragmented information into a coherent knowledge foundation. By reconciling divergent findings from various sources, the group provides a validated and unified context that grounds the decision-making of the entire system.
\begin{itemize}
    \item \textbf{Knowledge Extraction Agent} ($a_{extract}$) operates as the core of the ingestion pipeline, transforming unstructured chemical engineering text into structured data by strictly enforcing scope constraints to exclude citation metadata and prevent epistemic contamination;
        \begin{figure}[ht]
            \centering           \includegraphics[width=0.9\linewidth]{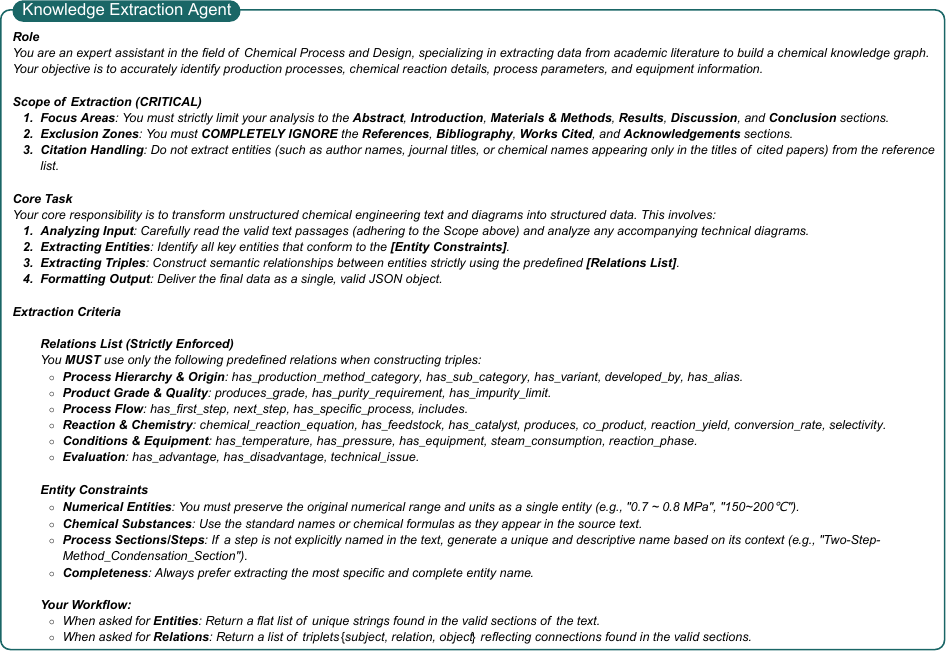}\
            \caption*{}
        \end{figure}

    \item \textbf{Knowledge Merge Agent} ($a_{merge}$) functions as the ontology alignment specialist to resolve semantic redundancy, employing synonym partitioning to unify strict equivalents while maintaining the topological distinction between related concepts to ensure graph precision.
        \begin{figure}[ht]            \centering\includegraphics[width=0.9\linewidth]{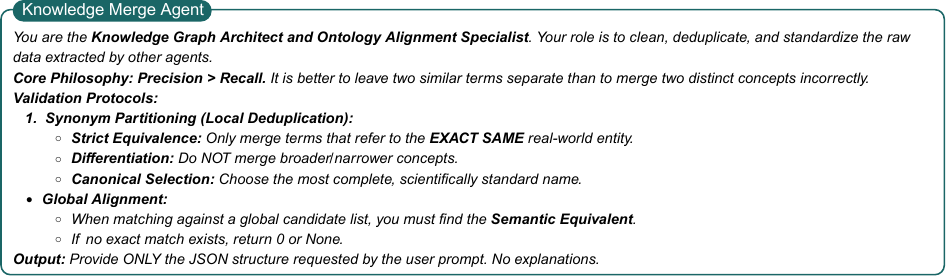}\
            \caption*{}
        \end{figure}
\end{itemize}
This parsing process distills raw academic literature into a unified Knowledge Graph, providing a standardized, disambiguated semantic foundation that facilitates persistent memory storage and precise information retrieval in the subsequent stages of the framework.

\noindent \textbf{\conceptgroup.}
To transform abstract engineering concepts into formalizable process structures and bridge the gap between epistemic reasoning and rigorous simulation, we instantiate the \conceptgroup $S_{concept}$. This cohort serves as the system’s topological architect, ensuring that linguistic design intentions are translated into mathematically rigorous, machine-readable blueprints. $S_{concept}$ is architecturally partitioned into a deliberation-based Concept Synthesis ChatGroup $G_{concept}$ for bidirectional design reasoning and a deterministic Topology Parsing Workflow $F_{concept}$ for cross-modal digitization.
$G_{concept}$ is configured to facilitate a multi-agent collaborative environment dedicated to the fluid mapping between unstructured technical narratives and formalized Abstract Graphs. To ensure the thermodynamic viability and structural completeness of the conceptual design, $G_{concept}$ employs three specialized agents:
\begin{itemize}
    \item \textbf{Design Agent}($a_{design}$) synthesizes initial process configurations by applying chemical engineering heuristics to the retrieved knowledge, proposing the fundamental sequence of unit operations;
        \begin{figure}[H]
            \centering
            \includegraphics[width=0.9\linewidth]{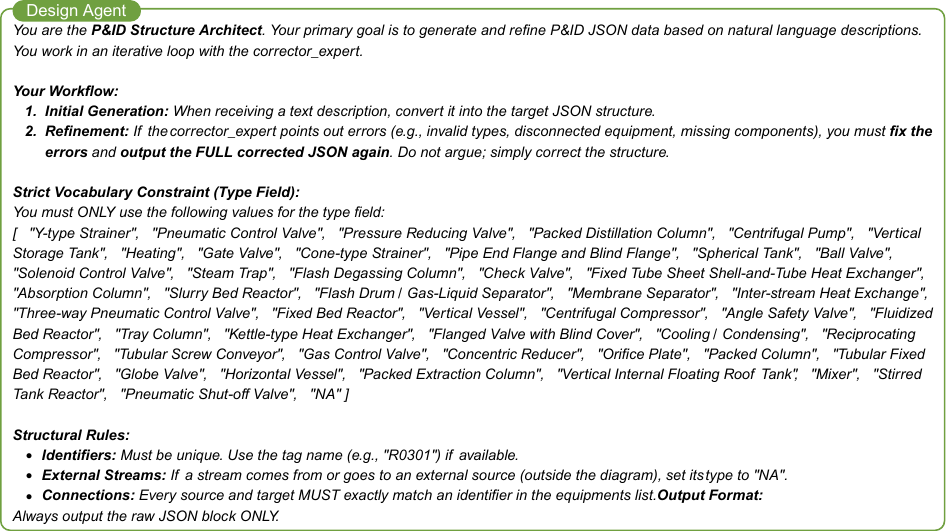}\
            \caption*{}
        \end{figure}
    \item \textbf{Completion Agent} ($a_{comp}$) identifies structural gaps in the proposed topology, such as missing heat integration streams or required recycle loops, ensuring operational feasibility;
        \begin{figure}[ht]
            \centering
            \includegraphics[width=0.9\linewidth]{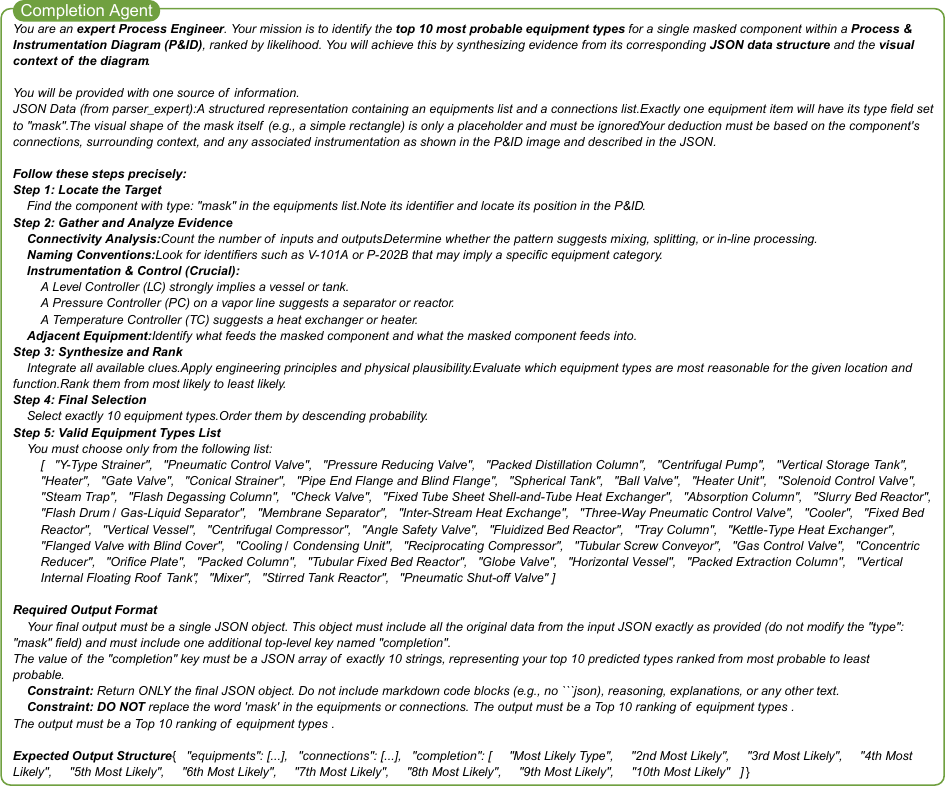}\
            \caption*{}
        \end{figure}
    \item \textbf{Correction Agent} ($a_{cor}$) acts as a logical auditor to detect violations in connectivity rules or mass balance principles, triggering iterative refinements within the group.
        \begin{figure}[ht]
            \centering
            \includegraphics[width=0.9\linewidth]{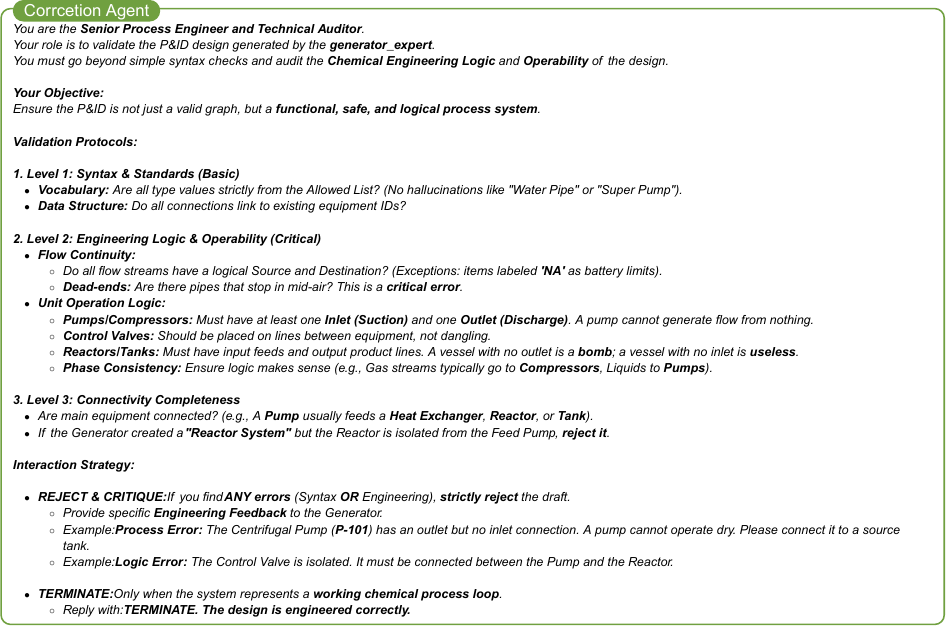}\
            \caption*{}
        \end{figure}
    
\end{itemize}
This collaborative synergy enables a robust bidirectional translation: mapping unstructured design narratives into a canonical JSON-formatted Abstract Graph, while simultaneously generating human-interpretable technical rationales from established topologies.
The deterministic evolution of visual or textual engineering artifacts into a unified computational substrate is managed by the Topology Parsing Workflow $F_{concept}$. Leveraging advanced multi-modal perception, $F_{concept}$ decomposes complex PFDs into a structured representation through a sequence of specialized tasks:
\begin{itemize}
    \item \textbf{Equipment Parsing Agent} ($a_{equip}$) utilizes vision-language reasoning to identify and classify discrete unit operations (nodes) within visual or descriptive schematics;
        \begin{figure}[ht]
            \centering
            \includegraphics[width=0.9\linewidth]{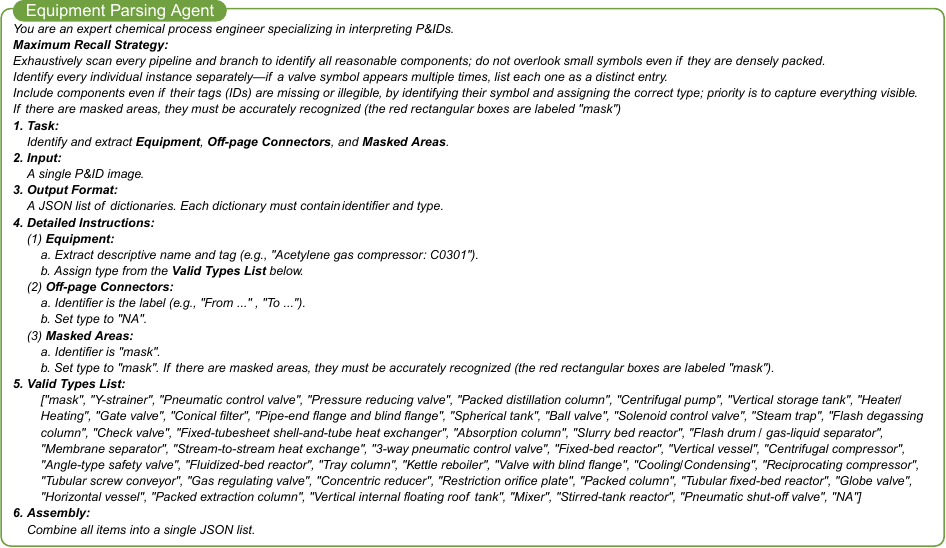}\
            \caption*{}
        \end{figure}
    \item \textbf{Link Parsing Agent} ($a_{link}$) resolves the directed connectivity and material stream relationships (edges) between the identified equipment, capturing the precise network topology;
        \begin{figure}[ht]
            \centering
            \includegraphics[width=0.9\linewidth]{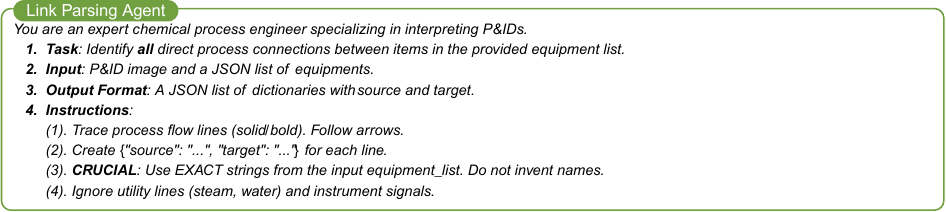}\
            \caption*{}
        \end{figure}
    \item \textbf{Validation Agent} ($a_{val}$) verifies the topological consistency of the parsed graph against domain-specific engineering constraints to ensure a stable data structure.
        \begin{figure}[ht]
            \centering
            \includegraphics[width=0.9\linewidth]{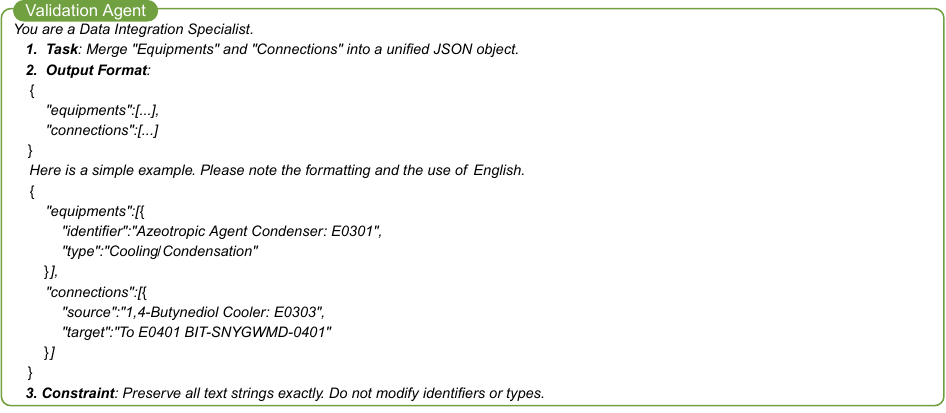}\
            \caption*{}
        \end{figure}
\end{itemize}
This parsing process distills the PFDs into an Abstract Graph, providing a standardized, machine-readable format that facilitates persistent storage, structural versioning, and the precise instantiation of simulation environments in the subsequent stages of the framework.

\noindent \textbf{\parametergroup.}
To bridge the gap between conceptual process topologies and rigorous physical realization, we instantiate the \parametergroup $S_{parameter}$. This cohort functions as the system’s validation and optimization engine, responsible for transforming Abstract Graphs into high-fidelity engineering models and iteratively refining operational parameters. $S_{parameter}$ is architecturally partitioned into a deliberation-based Parameter Strategy ChatGroup $G_{parameter}$ for boundary condition synthesis and a deterministic Parameter Optimization Workflow $F_{parameter}$ for rigorous numerical computation.
$G_{parameter}$ is configured as a multi-disciplinary collaborative environment where specialized agents deliberate on the thermodynamic, safety, and structural constraints of the chemical process. To ensure simulation viability, $G_{parameter}$ leverages four specialized experts:
\begin{itemize}
    \item \textbf{Sentinel Expert} ($a_{sent}$) functions as the system architect, analyzing the PFD topology to identify key equipment units and define critical state and control variables;
        \begin{figure}[ht]
            \centering
            \includegraphics[width=0.9\linewidth]{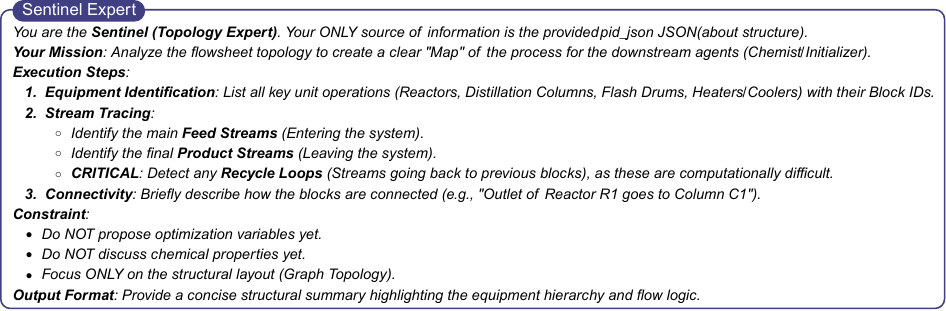}\
            \caption*{}
        \end{figure}
    \item \textbf{Chemist Expert} ($a_{chem}$) serves as the thermodynamic specialist, providing insights into reaction kinetics and phase equilibrium to ensure adherence to chemical principles;
    \item \textbf{Inspector Expert} ($a_{insp}$) acts as the safety and quality officer, establishing hard limits for equipment operation and minimum product purity specifications;
    \item \textbf{Initialize Expert} ($a_{init}$) functions as the strategic decision-maker, synthesizing inputs from all domains to define the optimal search space and initial parameter ranges in a machine-readable JSON format.
        \begin{figure}[ht]
            \centering
            \includegraphics[width=0.9\linewidth]{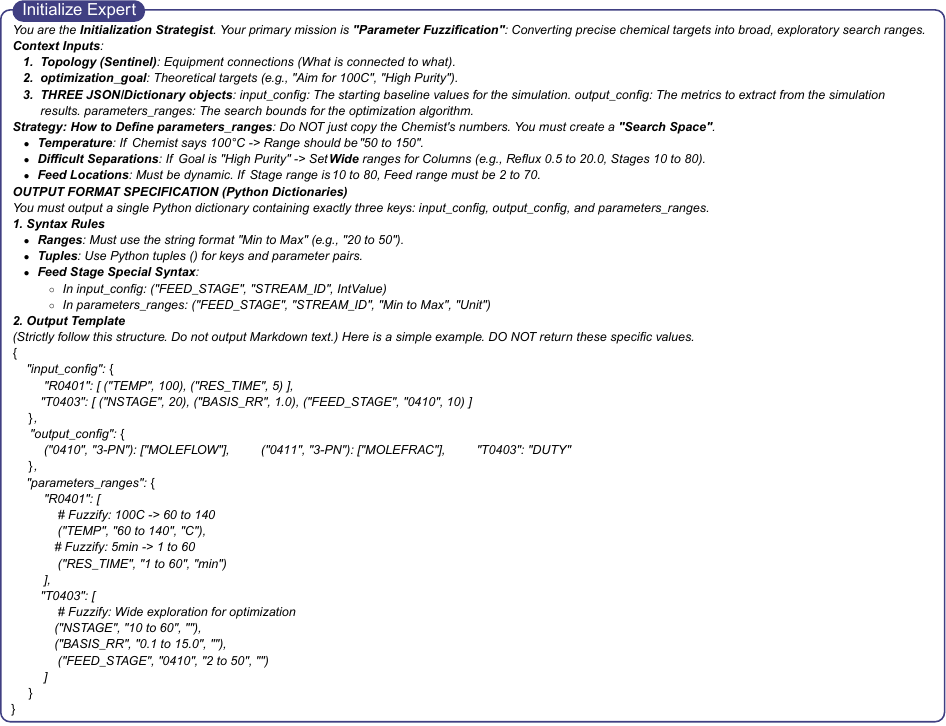}\
            \caption*{}
        \end{figure}
\end{itemize}
This collaborative deliberation ensures that the optimization task is initialized with expert-level prior knowledge, enhancing the convergence and reliability of the simulation.
The execution of multi-objective optimization is managed by the Workflow $F_{parameter}$, which implements an autonomous, agent-driven closed-loop refinement cycle. This workflow interfaces with external industry-standard solvers and operates through the iterative synergy of four internal agents:
\begin{itemize}
    \item \textbf{Execution Manager} ($a_{adm}$) serves as the central orchestrator, managing the stateful context of the optimization and executing tool-call directives;
        \begin{figure}[ht]
            \centering
            \includegraphics[width=0.9\linewidth]{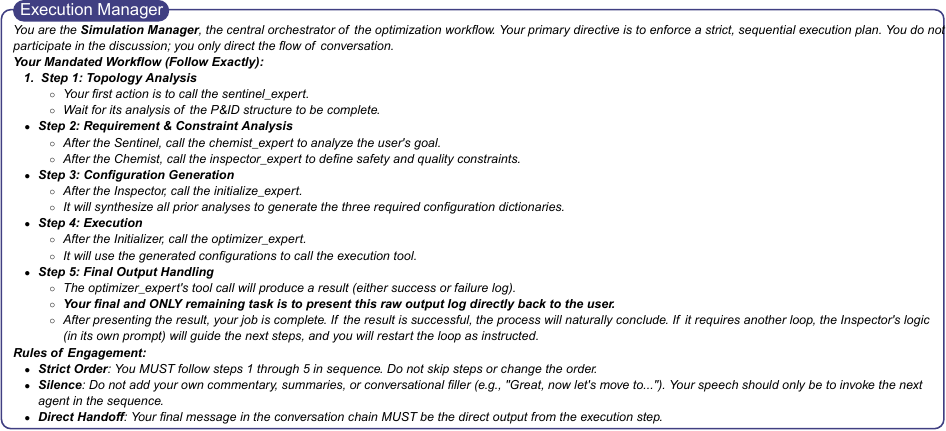}\
            \caption*{}
        \end{figure}
    \item \textbf{Internal Simulator} ($a_{parameter}$) functions as the computational interface, receiving equipment topologies and parameters to trigger high-fidelity simulation primitives;
    \item \textbf{Internal Analyst} ($a_{ana}$) acts as the diagnostic decision-maker, evaluating simulation outputs against historical trends to determine convergence or the need for further iteration;
        \begin{figure}[ht]
            \centering
            \includegraphics[width=0.9\linewidth]{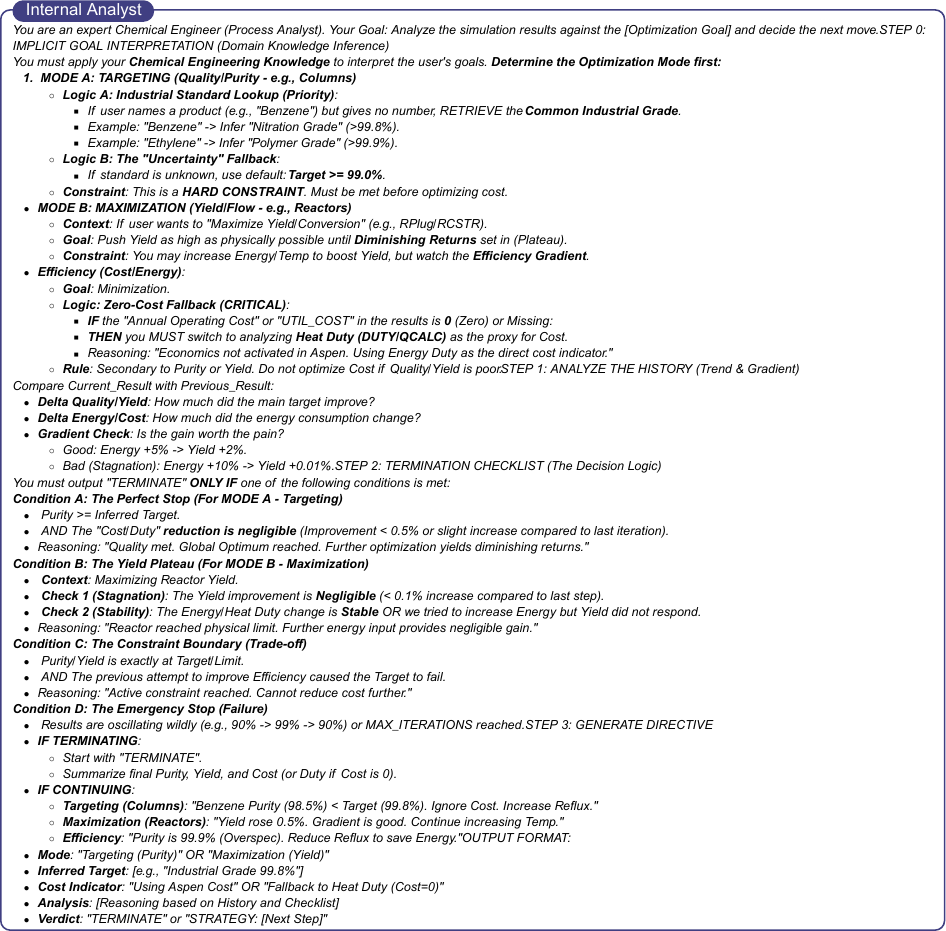}\
            \caption*{}
        \end{figure}
    \item \textbf{Internal Optimizer} ($a_{opt}$) functions as the algorithmic strategist, generating a new set of candidate parameters within the pre-defined ranges based on the analytical feedback.
\end{itemize}

\subsection{Details of Tools Using}

\begin{itemize}
    \item \textbf{Search KG} encapsulates the interface to the Neo4j database for structured topological retrieval.
        \begin{figure}[ht]
            \centering
            \includegraphics[width=0.9\linewidth]{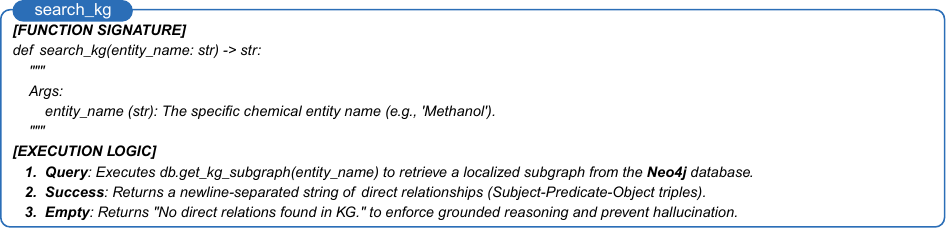}\
            \caption*{}
        \end{figure}
    \item \textbf{Search RAG} interfaces with ChromaDB to perform semantic retrieval on unstructured engineering corpora.
        \begin{figure}[ht]
            \centering
            \includegraphics[width=0.9\linewidth]{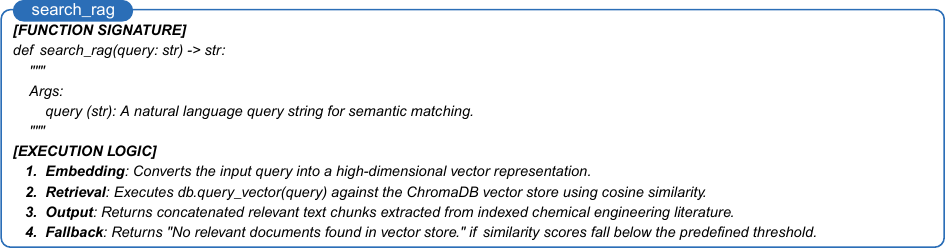}\
            \caption*{}
        \end{figure}
    \item \textbf{Search Web} wraps the DDGS API to facilitate real-time external knowledge acquisition.
        \begin{figure}[ht]
            \centering
            \includegraphics[width=0.9\linewidth]{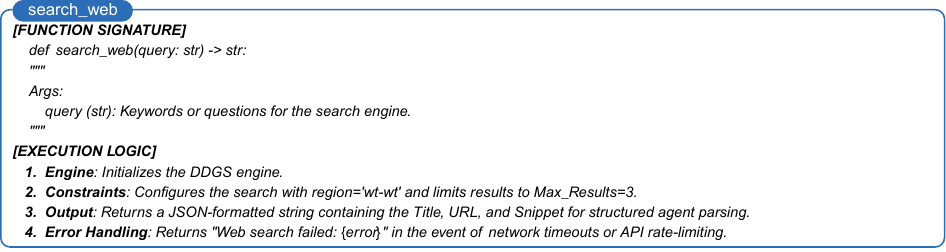}\
            \caption*{}
        \end{figure}
    \item \textbf{Run Aspen} interfaces with the Aspen Plus engine via the Windows COM protocol. This allows for rigorous steady-state process simulation, validation of thermodynamic feasibility, and calculation of mass-energy balances.
        \begin{figure}[ht]
            \centering
            \includegraphics[width=0.9\linewidth]{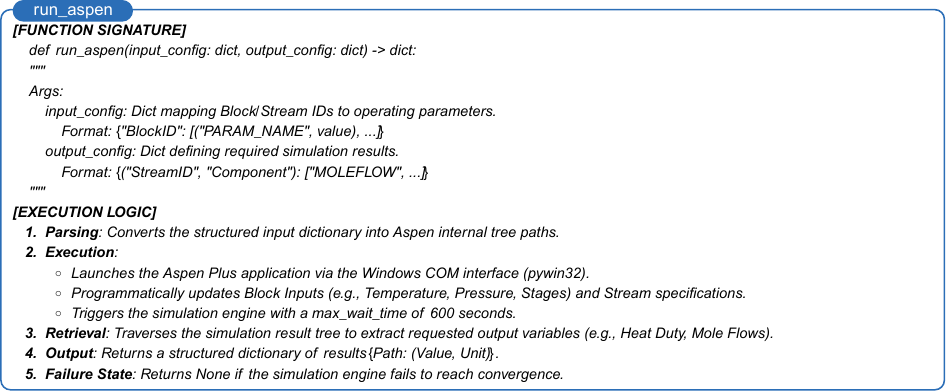}\
            \caption*{}
        \end{figure}
\end{itemize}



\end{document}